\documentclass[a4paper,fleqn]{cas-dc}
\usepackage[numbers]{natbib}
\usepackage{array}
\usepackage{longtable}   
\usepackage{pdflscape}   
\usepackage{caption}     
\usepackage{graphicx}

\DeclareCaptionFormat{myformat}{
  \textbf{#1} #3 }
\def\tsc#1{\csdef{#1}{\textsc{\lowercase{#1}}\xspace}}
\tsc{WGM}
\tsc{QE}
\tsc{EP}
\tsc{PMS}
\tsc{BEC}
\tsc{DE}

\begin{document}
\let\WriteBookmarks\relax
\def\floatpagepagefraction{1}
\def\textpagefraction{.001}

\shorttitle{Medical LLMs}

\shortauthors{J. Wang et~al.}

\title [mode = title]{A Survey on Large Language Models from General Purpose to Medical Applications: Datasets, Methodologies, and Evaluations}                      



%
\author[1]{Jinqiang Wang}[orcid=0000-0002-3711-0743]
\credit{Conceptualization, Methodology, Investigation, Writing
– original draft}


\ead{jqwang@xs.ustb.edu.cn}



\affiliation[1]{organization={School of Computer \& Communication Engineering, University of Science and Technology Beijing},
    city={Beijing},
    postcode={100083}, 
    country={China}}

\author[1]{Huansheng Ning}[orcid=0000-0001-6413-193X] \cormark[1]
\ead{ninghuansheng@ustb.edu.cn}
\credit{Conceptualization, Supervision}
\author[1]{Yi Peng}[orcid=0000-0003-2476-5908] \ead{ypeng123@foxmail.com, 2016020219@s.upc.edu.cn}
\credit{Conceptualization, Investigation}
\author[1]{Qikai Wei}[orcid=0000-0002-1048-3814] \ead{weiqikai@xs.ustb.edu.cn}
\credit{Validation, Writing – review \& editing}
\author[1]{Daniel Tesfai} \ead{danieltesfai353@gmail.com}
\credit{Validation, Writing – polish}
\author[1]{Wenwei Mao}[orcid=0009-0007-1902-7245] \ead{maowenwei@xs.ustb.edu.cn}
\credit{Validation, Writing – review \& editing}
\author[2]{Tao Zhu}[orcid=0000-0002-5879-5980] \ead{tzhu@usc.edu.cn}
\credit{Conceptualization, Validation, Supervision}
\author[3]{Runhe Huang}[orcid=0000-0001-5742-3766] \ead{rhuang@hosei.ac.jp}
\credit{Conceptualization, Supervision}


\affiliation[2]{organization={School of Computer Science, University of South China},
    city={Hengyang},
    postcode={421001}, 
    country={China}}


\affiliation[3]{organization={Faculty of Computer and Information Sciences, Hosei University},
    postcode={102-8160}, 
    country={Japan}}

\cortext[cor1]{Corresponding author}



\begin{abstract}
Large Language Models (LLMs) have demonstrated surprising performance across various natural language processing tasks. Recently, medical LLMs enhanced with domain-specific knowledge have exhibited excellent capabilities in medical consultation and diagnosis. These models can smoothly simulate doctor-patient dialogues and provide professional medical advice. Most medical LLMs are developed through continued training of open-source general LLMs, which require significantly fewer computational resources than training LLMs from scratch. Additionally, this approach offers better patient privacy protection than API-based solutions. Given the above advantages, this survey systematically summarizes how to train medical LLMs based on open-source general LLMs from a more fine-grained perspective. It covers (a) how to acquire training corpus and construct customized medical training sets, (b) how to choose an appropriate training paradigm, (c) how to choose a suitable evaluation benchmark, and (d) existing challenges and promising research directions are discussed. This survey can provide guidance for the development of LLMs focused on various medical applications, such as medical education, diagnostic planning, and clinical assistants. Related resources and supplemental information can be found on the GitHub repository \footnote{https://github.com/jqwangai/Medical-LLM}.
\end{abstract}
\fntext[fn1]{https://github.com/jqwangai/Medical-LLM}

\begin{keywords}
Large Language Model \sep Domain \sep Medicine \sep Healthcare
\end{keywords}

\maketitle

\section{Introduction}

Recently, large language models (LLMs), such as ChatGPT \citep{gpt-4}, have demonstrated exceptional capabilities in handling natural language processing (NLP) tasks. Their performance improves significantly as the model scaling increases while the size of the dataset keeps expanding. Currently, LLMs are widely applied in intelligent education, legal consultation, code generation, healthcare, and finance, showcasing their powerful NLP capabilities and broad applicability. In intelligent education \citep{dan2023educhat, wang2024large}, LLMs find application in intelligent tutoring systems and personalized learning planning, providing personalized advice by analyzing students' learning behaviors and feedback. In legal consulting \citep{cui2023chatlaw, colombo2024saullm}, LLMs facilitate legal document analysis, contract review, and legal question answering, improving the efficiency and accuracy of legal services. In code generation \citep{zhang2023unifying, du2024codegrag}, LLMs can automatically generate code snippets and perform code completions and fixes, augmenting development efficiency and code quality. In healthcare \citep{huatuogpt,InMD-X}, LLMs are used for medical literature analysis and doctor-patient dialogues, assisting medical decisions and improving patient care. In finance \citep{li2023large, lee2024survey}, LLMs are employed for risk assessment, financial analysis, and customer service, enhancing the intelligence and personalization of financial services.
\par
The success of these above domain LLMs relies on inheriting the strong knowledge of general LLMs and performing further training injected with domain-specific knowledge. The general LLM has acquired extensive knowledge, linguistic styles, and generalization abilities across a diverse corpus, enabling it to learn new tasks quickly. Therefore, the strategy of training domain LLMs based on general LLMs is more economical and effective than training domain LLMs from scratch \citep{ling2023beyond}.
There are three optional training stages for training the domain LLM from the general LLM \citep{zhao2023survey, ouyang2022training, gururangan2020don}: Continued Pretraining (CP), Instruction Fine-tuning (IFT), and Human Alignment (HA). CP involves training on unstructured domain data to help LLMs learn domain-specific knowledge, terminology, and language style. IFT focuses on training with domain instruction data, enabling LLMs to master domain-specific dialogue and instruction following. HA trains LLMs on human preference data, equipping them with domain-specific qualities such as harmlessness and helpfulness.
\par
Health constitutes the cornerstone of human survival, and advancements in medical technology significantly impact the quality of life and longevity \citep{zhang2020mental, RAHMATI2024128532, zhang2022internet}. Given the aging population and rising prevalence of chronic diseases, the medical domain confronts disparities in resource allocation, necessitating more efficient solutions \citep{TAO2024121699, chen2023digital}. \textbf{Medical LLMs can address this challenge by integrating rich medical data and clinical cases to help physicians quickly and accurately make diagnoses and formulate treatment plans}  \citep{thirunavukarasu2023large}. Furthermore, it can help medical institutions optimize resource allocation and improve the efficiency and quality of medical services. Existing medical LLMs focusing on different medical disciplines, such as internal medicine, respiratory medicine, and gastroenterology, can provide more specialized answers than general LLMs. Additionally, in medical consultation tasks, medical LLMs excel in interactive diagnosis, proactively prompting and guiding patients to provide missing information for ambiguous descriptions, distinguishing them from general LLMs.
\begin{figure*}[!htbp]
    \centering
    \includegraphics[width=1\linewidth]{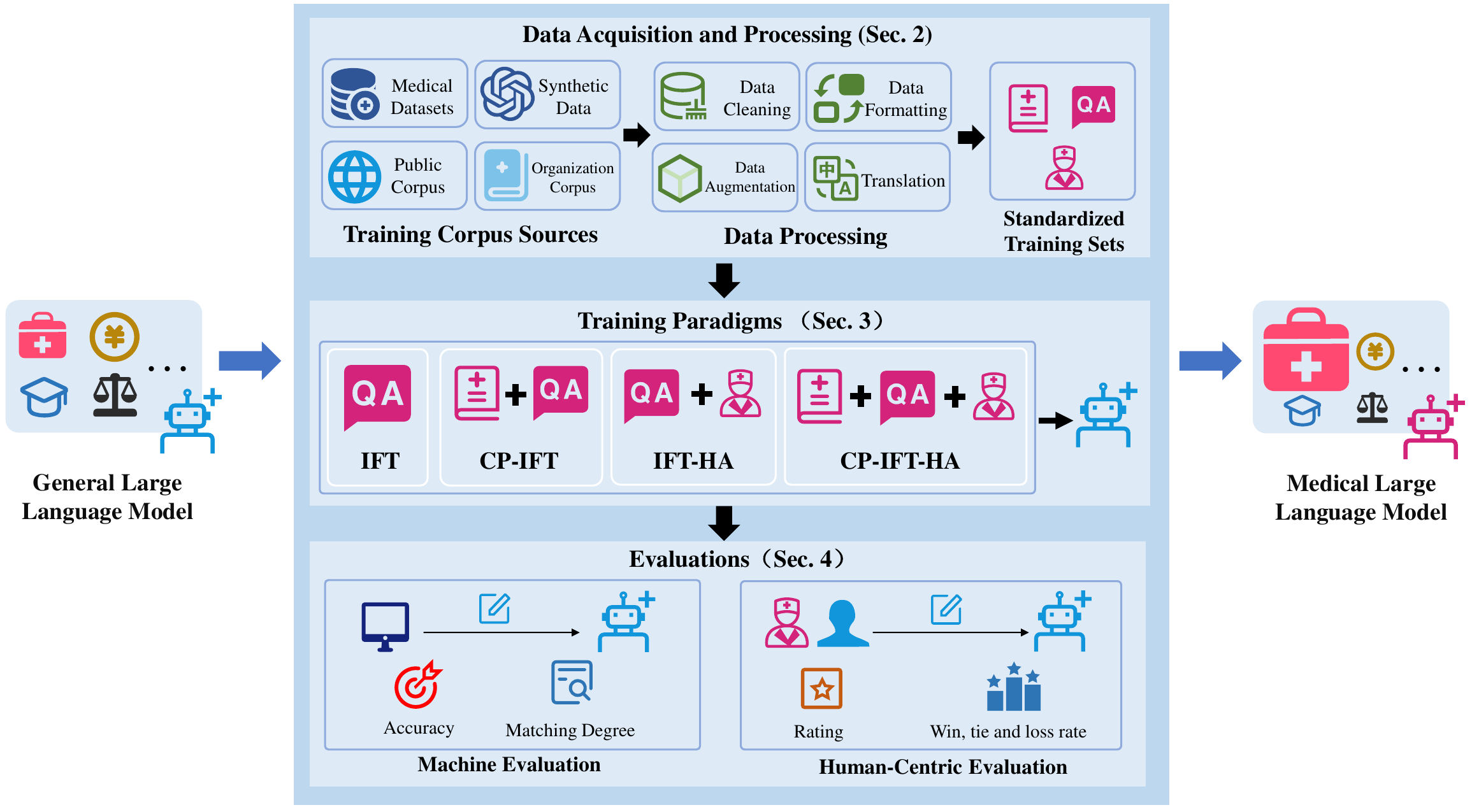}
    \caption{Training Pipeline from General LLMs to Medical LLMs. Firstly, the medical corpus is collected and processed to form a standardized training set. Next, an appropriate training paradigm is selected to train General LLMs to become Medical LLMs with medical knowledge. The training paradigms consist of three optional training stages: Continued Pretraining (CP), Instruction Fine-tuning (IFT), and Human Alignment (HA). Finally, Medical LLMs are evaluated from both machine and human perspectives. }
    \label{fig:overall}
\end{figure*}
\par
Given the significance of the healthcare domain, the advantages of medical LLMs over general LLMs, and the cost-effectiveness and efficiency of training domain LLMs from general LLMs, \textbf{this paper systematically surveys medical LLMs derived from further training of general LLMs from the perspectives of dataset, methodology, and evaluation.} The main content is depicted in Fig. \ref{fig:overall}. \textbf{Firstly}, we focus on the construction of training datasets for medical LLMs, as the quality of these datasets directly impacts the models' medical capabilities. Thus, we analyze the training datasets for existing medical LLMs from two perspectives: the source of the training corpus and data processing methods. Training corpus sources can assist developers in collecting the expected data, while data processing methods facilitate the cleansing of data to form a standardized training set. \textbf{Furthermore}, considering the differing capabilities injected into medical LLMs by the three training stages of CP, IFT, and HA, as well as the variations in required computational resources and data scale, we summarized the combinations of training stages adopted by existing medical LLMs and classified them into four paradigms: IFT, CP-IFT, IFT-HA, and CP-IFT-HA. These paradigms can guide institutions in choosing the appropriate training methods based on their computational capabilities, data scale, and specific needs. \textbf{Subsequently}, given the professional nature of content generated by medical LLMs and the potential harm that erroneous content can cause to patients, we systematically review existing evaluation benchmarks and protocols from two perspectives: machine evaluation and human-centric evaluation. In particular, we standardize the dimensions of the human-centric evaluation. \textbf{Finally}, we analyze the shortcomings of existing medical LLMs and propose future research directions with the potential to address these issues.
\par
Here, we compare several recent similar surveys. The survey \citep{he2023survey} systematically summarizes the training data, methods, optimization strategies, and evaluation techniques for LLMs in healthcare. However, it lacks sufficient depth in discussing data sources and evaluation techniques, and overlooks the summary of data processing methods. Compared to it, our work retrieves the latest literature and analyzes the training data from the perspective of 16 corpus sources. Additionally, our work analyzes data processing methods and evaluation techniques at a more fine-grained level. Another study similar to ours is \citep{he2024foundation}, which systematically overviews the methods, datasets, and application details of language foundation models, vision foundation models, bioinformatics foundation models, and multimodal foundation models in healthcare. However, it lacks an in-depth analysis of both the data, training strategies, and evaluation techniques for the language foundation models. Some other surveys \citep{thirunavukarasu2023large, busch2024systematic} focused on the applications and challenges of the LLMs in medicine, rather than on methods and technology.

\par
To bridge these gaps, the survey makes the following contributions:
\begin{enumerate}
    \item We systematically survey medical LLMs that emerged since ChatGPT. Detailed guidelines and tutorials are provided for healthcare researchers in training customized medical LLMs in terms of dataset construction, training paradigms, and evaluation technologies.
    \item Data Acquisition and Processing (Section \ref{sec:dataset}). We summarize and classify 16 training corpus sources and 4 groups of data processing methods for medical LLMs, providing robust recommendations and guidance for constructing customized medical datasets.
    \item Training Paradigms (Section \ref{sec:paradigms}). We systematically summarize the training stages for existing medical LLMs and classify them into 4 training paradigms. This categorization facilitates medical researchers in selecting appropriate training paradigms based on their available computational resources.
    \item Medical LLMs Evaluations (Section \ref{sec:evaluations}). We summarize the existing benchmark types for evaluating medical LLMs and categorize them into machine and human-centric evaluations. Additionally, the human-centric evaluation dimensions of medical LLMs are normalized. This result provides a comprehensive evaluation perspective and methodology for healthcare researchers.
    \item Challenges and Future Directions  (Section \ref{sec:challenges}). We propose future research directions for addressing the shortcomings and research gaps of existing medical LLMs. These directions provide important guidance and insights for medical organizations and researchers to explore medical LLMs further.
\end{enumerate}

\section{Data Acquisition and Processing} \label{sec:dataset}

\begin{figure}
    \centering
    \includegraphics[width=1\linewidth]{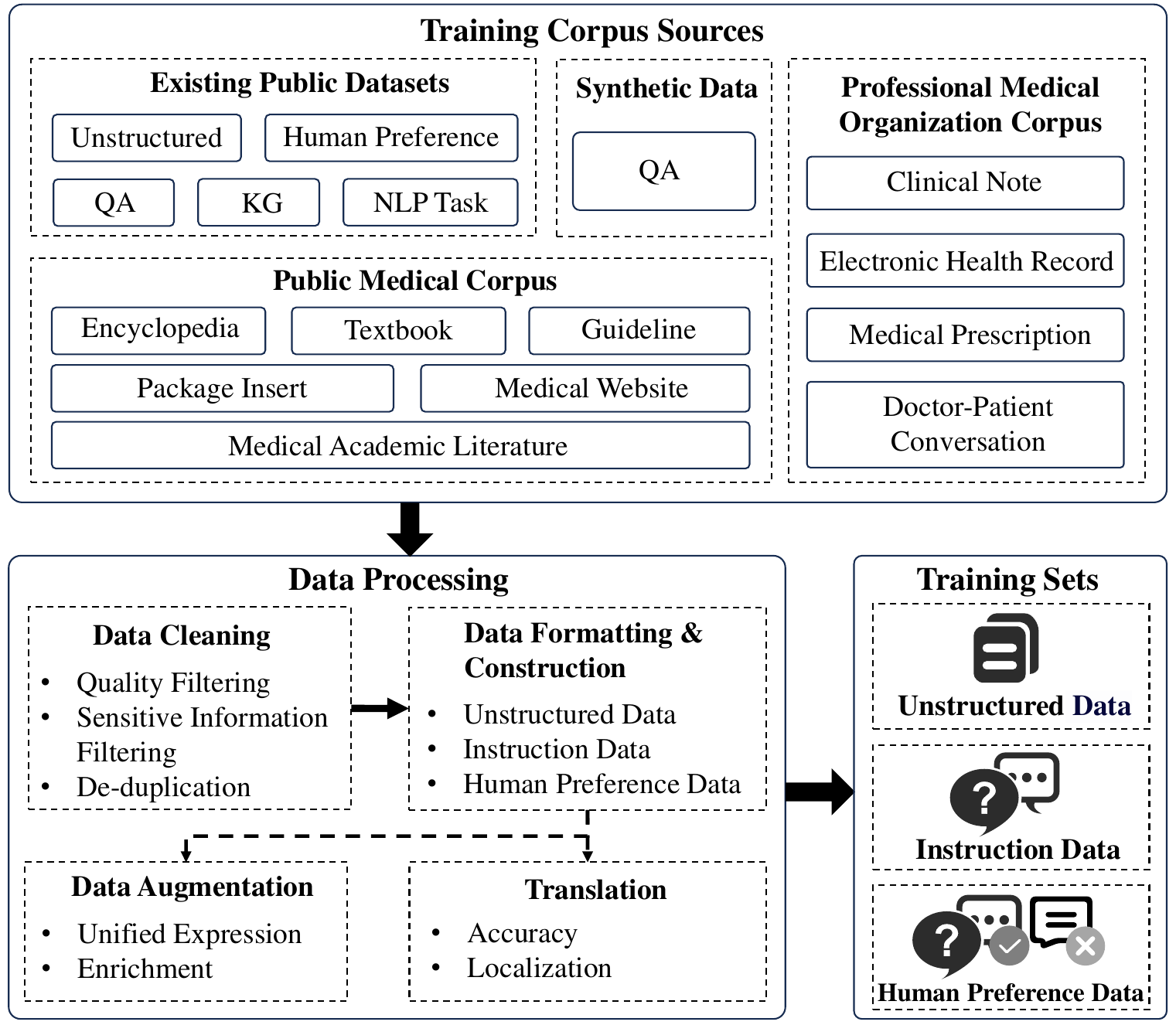}
    \caption{Detailed Categorization of Corpus Sources and Data Processing Methods.  }
    \label{fig:dataset}
\end{figure}

Data are crucial for transferring knowledge of LLMs from general purpose to domain applications. The quality of a dataset significantly impacts model performance. A high-quality dataset provides accurate supervised signals, enabling the model to effectively learn domain-specific rules and patterns. Moreover, this high-quality dataset not only improves the model's alignment with the target task but also breaks the scaling law to some extent \citep{gunasekar2023textbooks}. Medical LLMs should be doctor-like, patient-friendly, and professional \citep{huatuogpt}. To ensure that general LLMs are able to learn the language characteristics and terminology of the medical domain, training datasets should meet the above guidelines. To construct a medical domain dataset that meets the requirements of a specific task, it is essential to identify appropriate \textbf{training corpus sources}. These sources may vary in their medical discipline focus, level of specialization, and data quality. After identifying suitable corpus sources, the corpus cannot be directly used for model training. The data may contain inappropriate content, such as violent, pornographic, discriminatory material, and irregular data formats. Therefore, \textbf{data processing} is required to ensure the corpus meets the necessary training requirements.

The training of medical LLMs requires three types of data: unstructured data for continued pretraining, question-answer pairs (both single and multiple turns) for instruction fine-tuning, and human preference data for human alignment. Here, we systematically summarize the training corpus sources and data processing methods for investigated medical LLMs in Tab. \ref{tab:dataset}, providing a reference for constructing custom medical training sets. In addition, we categorized the corpus sources and data processing methods, as shown in Fig. \ref{fig:dataset}.

\subsection{Training Corpus Sources}
\begin{figure}
    \centering
    \includegraphics[width=1\linewidth]{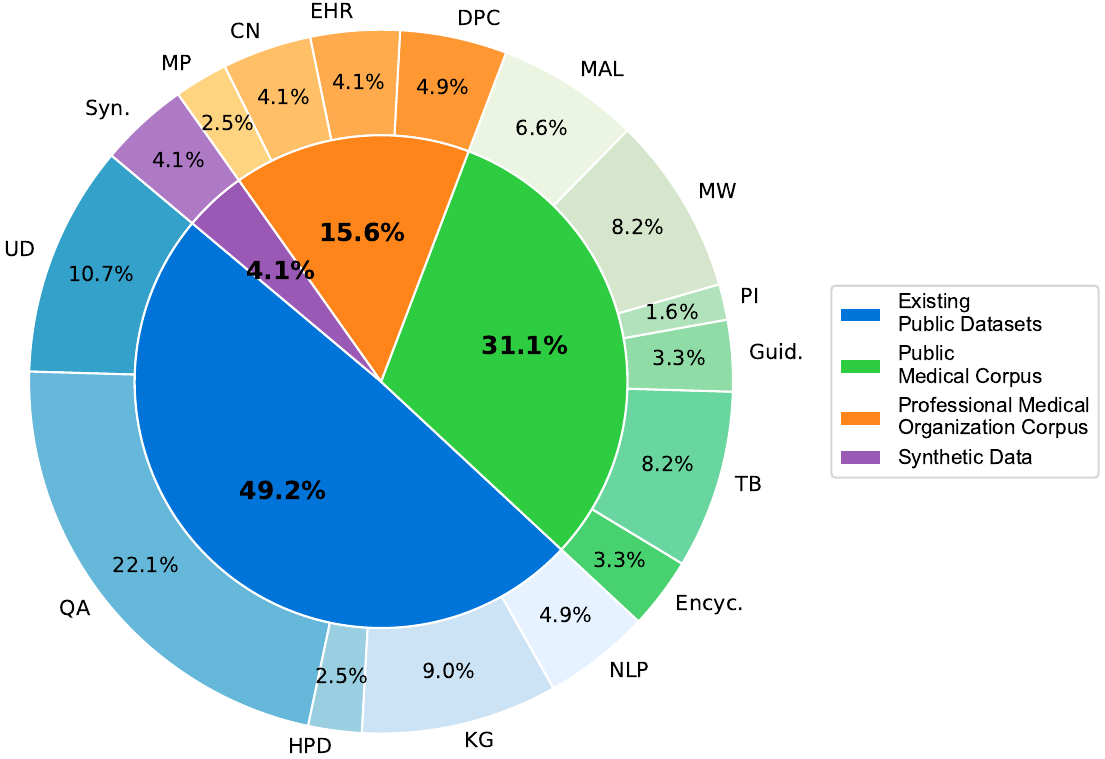}
    \caption{Percentage of frequency for each corpus source. The abbreviations are the same as in Tab. \ref{tab:dataset}.}
    \label{fig:sources}
\end{figure}

We categorized the corpus sources into 4 main groups: Existing Public Datasets, Public Medical Corpus, Professional Medical Organization Corpus, and Synthetic Data. These groups were further subdivided into 16 categories, as illustrated in Fig. \ref{fig:dataset}.
The strengths and weaknesses of each category, as well as their suitability for different training stages of medical LLMs, are detailed below. Additionally, the guidance is provided for the collection of customizing training data. Finally, we counted the frequency of each corpus source according to the data in Tab. \ref{tab:dataset}, as shown in Fig. \ref{fig:sources}.

\onecolumn
\begin{landscape}
\begin{longtable}{m{3cm}|>{\raggedright\arraybackslash}m{2cm}|>{\raggedright\arraybackslash}m{1.8cm}|>{\raggedright\arraybackslash}m{1.8cm}|m{0.5cm}|c|c|c|c|m{2cm}|m{2cm}|m{2cm}} 
\caption{\\Corpus Sources, Processing Methods, Training Set Format and Scale for Medical LLMs. The abbreviations here are Pub. Data: Existing Public Datasets, Pub. Corpus: Public Medical Corpus, Org. Corpus: Professional Medical Organization Corpus, Syn.:Synthetic Data, clean.: Data Cleaning, Format.: Data Formatting \& Construction, Aug.: Data Augmentation, Trans.:Translation, QA: Question-Answer datasets (single turn or multiple turns or multiple choice), UD: Unstructured Data, InsD: Instruction Data, HPD: Human Preference Data, KG: Knowledge Graph, NLP: NLP Task, MB: Medical Website, Encyc.: Encyclopedia, TB: Textbook, Guid.: Guideline, PI: Package Insert, MAC: Medical Academic Literature, CN: Clinical Note, EHR: Electronic Health Record, MP: Medical Prescription, DPC: Doctor-Patient Conversation.} \\  
\toprule
\label{tab:dataset}
\multirow{2}{*}{Models} & \multicolumn{4}{c}{Training Corpus Sources}  & \multicolumn{4}{c}{Data Processing} & \multicolumn{3}{c}{Standardized Training Sets (Size)} \\ 
\cmidrule(r){2-5}\cmidrule(r){6-9} \cmidrule(r){10-12}
& Pub. Data.  & Pub. Corpus  & Org. Corpus & Syn. & Clean. & Format. & Aug. & Trans. & UD &InsD &HPD  \\
\hline
\endfirsthead
\captionsetup{justification=raggedright, singlelinecheck=false}
\caption{(continued)} \\ 
\hline
\multirow{2}{*}{Models} & \multicolumn{4}{c}{Training Corpus Sources}  & \multicolumn{4}{c}{Data Processing} & \multicolumn{3}{c}{Standardized Training Sets} \\ 
\cmidrule(r){2-5}\cmidrule(r){6-9} \cmidrule(r){10-12}
& Pub. Data.  & Pub. Corpus  & Org. Corpus & Syn. & Clean. & Format. & Aug. & Trans. & UD &InsD &HPD  \\
\hline
\endhead

Med-PaLM \citep{med-palm}               & QA                       &                MW & &                                          &               &                 &                   &            & & 65 items &\\ \hline 
ChatDoctor \citep{chatdoctor}              &                          &                MW & &                                          & $\checkmark$             &                 &                   &             && 100K items&\\ \hline
DoctorGLM  \citep{doctorglm}             &                         &               &                      &                                         &             &                &                  & $\checkmark$            & &4,487K items &\\  \hline 
BenTsao  \citep{bentsao}      &  KG                       &               &                      &                                         &             &                $\checkmark$&                  & & & 8K items &\\ \hline 
ChatGLM-Med \citep{ChatGLM-Med}
& KG                       & & & & & $\checkmark$& & &&7.6K items&\\ \hline 
MedAlpaca \citep{MedAlpaca}
& QA, NLP& MW & & & $\checkmark$& $\checkmark$& & & & 248K items &\\ \hline 
PMC-LLaMA \citep{PMC-LLaMA}
& UD, QA, KG& MAL, TB& & & $\checkmark$& $\checkmark$& $\checkmark$& & 79B tokens & 202M tokens &  \\ \hline 
HuatuoGPT \citep{huatuogpt}
& QA& & DPC& $\checkmark$& $\checkmark$& & $\checkmark$& & & 226K items &\\ \hline 
ChatMed-Consult \citep{ChatMed-Consult}& & MW & & $\checkmark$& & $\checkmark$& & & &110K items & \\ \hline 
Med-PaLM 2 \citep{med-palm-2}& QA& & & & & & & &&193K items&\\ \hline 
Clinical Camel \citep{clinical-camel} & QA& MAL& & & & $\checkmark$& $\checkmark$& &&174K items&\\ \hline 
ShenNong-TCM \citep{ShenNong-TCM} & KG                       & & & & & $\checkmark$& & &&113K items&\\ \hline  
MedicalGPT \citep{MedicalGPT} & UD, QA, HPD& & & & & & & &361K items & 2.07M items& 3.8K items\\ \hline 
ClinicalGPT \citep{ClinicalGPT} & QA, KG& & EHR& & & & & &&423K items& 10K items\\ \hline 
DISC-MedLLM \citep{Disc-medllm}
& QA, KG& & & & & $\checkmark$& $\checkmark$& & &514K items&\\ \hline 
 Zhongjing \citep{zhongjing}
& QA, KG,  NLP& TB& EHR, CN, \newline DPC& & & $\checkmark$& $\checkmark$& & 1,086 MB & 470K items& 20K items\\ \hline 
  BianQue \citep{Bianque}
& & & DPC& & $\checkmark$& & $\checkmark$& & & 2,437K items &\\ \hline 
   Alpacare \citep{alpacare}
& & & & $\checkmark$& & $\checkmark$& & &&52K items&\\ \hline 
    Qilin-Med \citep{Qilin-med}
& QA, KG, \newline HPD& & & & & $\checkmark$& & &2B tokens&11M items& 6K items\\ \hline 
Taiyi \citep{Taiyi}
& QA, NLP& & & & $\checkmark$& $\checkmark$& & &&1,114K items&\\ \hline 
ChiMed-GPT \citep{ChiMed-GPT}
& UD, QA, HPD& & & & $\checkmark$& & $\checkmark$& &369K items & 1,269K items & 4K items\\ \hline 
MediTron \citep{Meditron}
& UD, QA& MAL& CN& & $\checkmark$& & $\checkmark$& &48 tokens&369K items&\\ \hline 
HuatuoGPT-II \citep{Huatuogpt-ii}
& UD& Encyc., TB, MAL, MW& & $\checkmark$& $\checkmark$& $\checkmark$& & $\checkmark$&&5,394K items&\\ \hline 
AntGLM-Med \citep{AntGLM-Med}& UD, QA, KG& TB, MAL& DPC& & $\checkmark$& $\checkmark$& & &15.39B tokens & 632K items &\\ \hline 
GPT-doctor \citep{GPT-doctor}
& & MW, PI& & & & $\checkmark$& & &&1,939K items&\\ \hline 
EpilepsyLLM \citep{EpilepsyLLM} & QA & MW &MP, CN & & & $\checkmark$ & & $\checkmark$ & & 52.2K items& \\ \hline
BioMistral \citep{BioMistral}
& UD, QA& & & & & & & &3B tokens&405K items&\\ \hline
MMedLM \citep{MMedLM}& UD& MW, TB& & & $\checkmark$& & & &25.5B tokens&&\\ \hline
InMD-X \citep{InMD-X}& & MAL& & & $\checkmark$& $\checkmark$& & &150M tokens & 1,701K items &\\ \hline
Me-LLaMA \citep{Me-LLaMA} & UD, QA, \newline KG, NLP& MAL& EHR, CN& & & $\checkmark$& & &129B tokens & 214K items&\\ \hline
JMLR \citep{JMLR} & QA& TB, Guid.& EHR& & & & & & - & - & \\  \hline
BiMediX \citep{BiMediX} & QA& & & & & $\checkmark$& & $\checkmark$& &1,311K items&\\  \hline
OncoGPT \citep{OncoGPT} & & MW& DPC& & $\checkmark$& & & &&332K items&\\  \hline
Apollo \citep{Apollo} & UD, QA& TB, MAL, MW, Encyc., Guid. & & & $\checkmark$& $\checkmark$& & & 2,054M tokens& 481M items&\\ \hline
Qibo \citep{Qibo} & QA, NLP& TB, Encyc.& MP& & $\checkmark$& & & &1,277MB & - &\\ \hline
Hippocrates \citep{hippocrates} & UD, QA & & & & &  $\checkmark$ & & & 298M tokens & 182.9M tokens & 15,258 items \\ \hline
MING-MOE \citep{MING-MOE} & QA, NLP & & & & &  $\checkmark$ & & & &300K tokens&\\ \hline
LingDan \citep{lingdan} & UD & PI, TB & MP, CN, EHR & &$\checkmark$ & $\checkmark$ & & & 304M tokens & 201.5K items & \\ \hline
Aloe \citep{aloe} & QA & Guid. & & $\checkmark$ & $\checkmark$ & $\checkmark$ & $\checkmark$ & & & 872K items& 12k items \\ \hline
PediatricsGPT \citep{PediatricsGPT} & QA. KG & TB, Encyc., Guil. & DPC & & $\checkmark$ & $\checkmark$ & & & 975.8MB & 332K items & 15,556 items \\ \hline
Aqulia-Med \citep{Aqulia-Med} & UD, QA & & & & $\checkmark$ & $\checkmark$ & $\checkmark$ & & 80B tokens & 320K items & 12,727 items \\
\bottomrule

\end{longtable}
\end{landscape}
\twocolumn

\subsubsection{Existing Public Datasets}
It is recommended that existing public datasets be used to construct customized training sets for medical LLMs. These datasets are well-designed, high-quality, and widely recognized in existing research. The public datasets are further categorized into five types: unstructured data, question-answer pairs, human preference data, knowledge graphs, and regular NLP task data. Related work \citep{he2023survey, he2024foundation} have systematically summarized datasets for LLMs in the medical domain. Here, we focus on the characteristics of each data type and considerations for using these public datasets.
\par 
\textbf{Unstructured Data} usually refers to plain text data, such as the numerous public corpora including CommonCrawl \citep{CommonCrawl}, C4 \citep{raffel2020exploring}, Wikipedia \citep{Wikipedia}, Baidu Baike \citep{Baidu_Baike}, UFAL Medical Corpus \citep{UFAL} , and other LLM collections like RedPajama \citep{together2023redpajama}. These corpora are collected from web data using crawlers, resulting in a large-scale dataset. However, they require rigorous data cleaning, such as filtering for medical-related content. The cleaned data can then be used to continued pretraining Medical LLMs, enhancing their understanding of the distributional and statistical characteristics of medical data and thereby improving the stability and robustness of the models.
\par
\textbf{Question-Answer Pairs (QA)} include single-turn question-answer, multi-turn question-answer (conversations), and multiple-choice question-answer datasets. Single-turn question-answer datasets, such as Huatuo-26M \citep{li2023huatuo} and ChiMed \citep{tian2019chimed}, primarily consist of questions and answers collected from real-life scenarios. Each question addresses a specific topic within the medical domain, enabling medical LLMs to learn a specialized manner and style of responding. 
However, due to its single-turn QA nature, it falls short of enhancing the ability of medical LLMs to process contextual information. Conversely, datasets in a multi-turn QA format, such as MedDialog \citep{zeng2020meddialog}, BianqueCorpus \citep{Bianque}, and CMtMedQA \citep{zhongjing}, have the potential to augment the capacity of medical LLMs to handle contextual information and simulate authentic doctor-patient interactions. Additionally, these multi-turn QA datasets often include instances where doctors inquire about patients' conditions, thereby bolstering the medical LLMs' ability to ask proactive questions and guide users in describing their situations. 
Multiple-choice QA datasets, such as MMLU \citep{hendrycks2020measuring}, MedQA \citep{jin2021disease}, PubMedQA \citep{jin2019pubmedqa}, and MedMCQA \citep{pal2022medmcqa}, consist of questions and options, with some derived from actual medical exam questions. In addition to correct answers, some of these datasets include detailed explanations. This enhances the model's ability to select correct answers and improves its understanding and reasoning about complex medical information.
\par
\textbf{Human Preference Data}, such as Zhongjing\_rlhf \citep{zhongjing} and MedicalGPT \citep{MedicalGPT}, consist of a question, a chosen response, and a rejected response. The chosen response aligns with human values (more like a doctor's response), whereas the rejected response is not consistent with human values compared to the chosen response. These datasets typically require manual labeling or verification, resulting in a smaller number of public datasets of this type compared to other types. Such datasets are used in the human alignment training stage of medical LLMs to ensure that the model's responses align with human linguistic conventions and exhibit characteristics that are doctor-like, patient-friendly, harmless, and professional.
\par
\textbf{Knowledge Graph} represents concepts, entities, and their interrelationships in the real world in a structured format, aligning information more closely with human cognition. It enhances the organization, management, and comprehension of vast data \citep{hogan2021knowledge, feng2022novel, ji2021survey, WANG2023120211}. 
Medical knowledge graphs usually contain rich structured medical knowledge with low data noise. This type of data can motivate medical LLMs to more accurately process medical terminology and causal relationships, thus improving models' performance in reasoning medical tasks. Common medical knowledge graphs include UMLS \citep{lindberg1993unified}, CMeKG \citep{byambasuren2019preliminary}, and BIOS\citep{BIOS}. Since these are not in QA pairs or plain text format, a carefully designed data transformation pipeline is required to convert them.
\par
\textbf{NLP Task} datasets include Named Entity Recognition (NER) \citep{li2020survey, navarro2023clinical}, Relation Extraction (RE) \citep{zhang2017review, raza2023entity}, Causal Relation Extraction (CRE) \citep{yang2022survey}, Text Classification (TC) \citep{minaee2021deep, ZHANG2024128617}, Semantic Textual Similarity (STS) \citep{wang2020medsts, YUAN2023126801}, Natural Language Inference (NLI) \citep{agrawal2019ars_nitk}, and others. These datasets enhance model performance in various medical tasks and instruction following capabilities. Me-LLaMA \citep{Me-LLaMA} and Taiyi \citep{Taiyi} utilize these conventional medical NLP datasets to improve their models' generalization capabilities. It is important to note that these data generally cannot be used directly for training medical LLMs and must be transformed into plain text or question-answer pair format by data processing.
\par
Public datasets are easily accessible and high quality, but they do not all meet the needs of training medical LLMs in the medical domain, such as focusing on pediatrics or other nuanced domains of medical LLMs. It is necessary to create an individual training set to meet customized needs. 

\subsubsection{Public Medical Corpus}
Publicly accessible online medical data is a primary source for training medical LLMs. The data encompasses diverse medical information from various regions, populations, and platforms. While it offers extensive coverage and a wide range of medical knowledge, it often presents challenges related to privacy and data quality, necessitating further cleaning and processing. Here, we summarize six common data sources for medical LLMs: medical encyclopedias, medical textbooks, medical academic literature, medical websites, medical guidelines, and package inserts.
\par
\textbf{Medical Encyclopedias} are comprehensive reference resources, typically available online, that systematically document a wide range of medical knowledge, including diseases, symptoms, diagnostic methods, treatments, and related medical technologies \citep{gellman2020encyclopedia}. Compiled by professional medical groups, academic institutions, or authoritative publishers, their content undergoes rigorous review and regular updates to ensure accuracy and currency. The structured organization of these encyclopedias facilitates efficient data retrieval and processing. Notable medical encyclopedia websites include MedlinePlus \citep{MedlinePlus}, Mayo Clinic \citep{Mayo_Clinic}, and WebMD \citep{WebMD}. Medical LLMs often use encyclopedias as a training data source to enhance their medical knowledge \citep{Huatuogpt-ii, Apollo}. Although the format of encyclopedia content is suitable for the pretraining of these models, further data processing and transformation are necessary for instruction fine-tuning.
\par
\textbf{Medical Textbooks} provide a systematic introduction to medical knowledge and skills, serving as comprehensive references for medical students, physicians, nurses, and other healthcare professionals \citep{tez2017reliable, kim2024small}. They cover a wide range of topics from basic sciences (e.g., anatomy, physiology, biochemistry) to clinical practice (e.g., internal medicine, surgery, pediatrics, obstetrics and gynecology). Authored by experts and peer-reviewed, these textbooks ensure the accuracy and authority of medical information. Common sources include FreeBooks4Doctors \citep{FreeBooks4Doctors}, AccessMedicine \citep{AccessMedicine}, and Bookshelf \citep{Bookshelf}. Similar to medical encyclopedias, medical textbooks offer extensive, specialized, and structured information. Consequently, medical LLMs \citep{AntGLM-Med, JMLR, Qibo, MMedLM} utilize these textbooks as training data to incorporate professional and precise medical knowledge. However, the specialized terminology and complex linguistic structures in medical textbooks can limit their accessibility for users without a medical background.
\par
\textbf{Medical Academic Literature} encompasses specialized publications that research, explore, and discuss various fields of medicine \citep{lemley2024does}. This literature includes peer-reviewed journal articles, scholarly books, conference papers, and research reports. Covering specialties such as clinical medicine, biomedical sciences, public health, pharmacology, and physiology, these publications provide up-to-date medical knowledge and evidence for researchers, clinicians, students, and allied professionals. Common sources include PubMed \citep{PubMed}, Embase \citep{Embase}, and ScienceDirect \citep{ScienceDirect}.
Due to its specialization, accuracy, and currency, medical academic literature is frequently used to train medical LLMs \citep{Meditron, Me-LLaMA,PMC-LLaMA}, enhancing their medical expertise. However, the specialized terminology and technical content may make these models produce responses that are not easily understood by the general public. Additionally, acquiring academic literature data requires consideration of copyright issues.
\par
\textbf{Medical Websites} encompass various online platforms, such as forums, consultations, news, and educational resources. Specifically, medical consultation platforms allow patients to communicate with doctors online for medical advice, diagnoses, and treatment plans, enhancing the convenience and accessibility of healthcare services. These platforms also offer doctors a flexible way to provide care more efficiently. Common medical websites include 98point6 \citep{98point6}, Ding Xiang Yuan \citep{Ding_Xiang_Yuan}, Teladoc Health \citep{TeladocHealth}, and Haodf \citep{Haodf}.
For training medical LLMs (LLMs), conversation data from medical consultation platforms is particularly valuable. This data is generally more accessible to the public compared to academic literature and textbooks due to its real-world, colloquial nature. Additionally, the multi-turn QA format helps improve the models' ability to process contextual information. Several models, such as MMedLM \citep{MMedLM}, ChatDoctor \citep{chatdoctor}, and GPT-Doctor \citep{GPT-doctor}, use this type of data to align their response styles with those of real doctors.
However, data from medical consultation platforms must be collected with strict adherence to privacy regulations to ensure patient information security. The quality of this data can vary, with some information being inaccurate or incomplete. Therefore, it is essential to screen and clean the data to maintain the accuracy and reliability of the medical information used for training.
\par
\textbf{Medical Guidelines}, also known as clinical or practice guidelines, are documents that provide recommendations for medical decision-making, including the diagnosis, management, and treatment of specific health conditions. These guidelines are critical for medical practice and are typically developed by professional medical organizations, government agencies, or academic groups \citep{berg1997clinical}. Common sources include CCO \citep{CCO}, CDC \citep{CDC}, NICE \citep{NICE}, and WHO \citep{WHO}.
Medical guidelines offer practical information across a broad spectrum of diseases and clinical scenarios, aiding medical LLMs in learning diagnostic and therapeutic decision-making. The standardized format and consensus-driven content of these guidelines enable models to assimilate best practices in the medical community. Models such as MediTron \citep{Meditron}, Apollo \citep{Apollo}, and JMLR \citep{JMLR} incorporate guideline data in their training, helping them provide practical advice to healthcare professionals, thus enhancing diagnostic accuracy and reducing the risk of misdiagnosis and mistreatment.
Despite their value, medical guidelines have limitations. They are often based on average data from the general patient population, which may not always be applicable to individualized clinical decision-making.
\par
\textbf{Package Inserts} are documents that provide comprehensive information about medications and their proper use \citep{fuchs2005survey}. Targeted at patients, doctors, and other healthcare professionals, it promotes the safe and effective use of medications. Package inserts typically include details such as the drug name and ingredients, usage and dosage, indications, contraindications, warnings and precautions, drug interactions, and side effects. Common sources for drug insert data include RxList \citep{RxList} and Drugs \citep{Drugs}.
These documents contain rich information that can enhance medical LLMs' understanding of drugs and support decision-making processes. However, there are limitations. Package inserts offer standardized medication instructions and may not account for individual patient conditions or concurrent medication use, which can affect the ability of medical LLMs to provide personalized healthcare and treatment recommendations. 
\par
Overall, the aforementioned public medical corpus sources offer a wealth of information and extensive coverage, providing diverse knowledge resources for medical LLMs. Additionally, they are relatively easy to access without significant time and resource investment. However, these public medical corpus sources also have some drawbacks. Firstly, due to the diversity of medical data sources, they need to be filtered in a fine-grained way to obtain data that matches the target task. Secondly, most of these data are unstructured and require appropriate data processing operations to convert them into the format required for instruction fine-tuning.

\subsubsection{Professional Medical Organization Corpus}
Data from specialized medical institutions, such as hospitals and healthcare centers, is a crucial information source for training medical LLMs. These data include patients' medical histories, diagnostic information, treatment plans, and doctor-patient interactions. Here, we summarize four commonly used professional medical organization corpus sources: Clinical Notes, Electronic Health Records, Medical Prescriptions, and Doctor-Patient Conversations.
\par
\textbf{Clinical Notes} are records created by healthcare professionals during the consultation, diagnosis, treatment, and monitoring of patients. Typically prepared by doctors, nurses, and therapists, these notes include patient information, symptom descriptions, diagnoses, assessments, treatment plans, and progress notes \citep{sivarajkumar2024mining}. MIMIC \citep{johnson2018mimic} is a publicly accessible database of clinical notes, offering diverse and comprehensive data covering the entire course of a patient's medical care.
These data will enhance medical LLMs' abilities in clinical decision-making and case understanding. For instance, models like Zhongjing \citep{zhongjing} and Me-LLaMA \citep{Me-LLaMA} use clinical notes for continued pretraining to enrich their knowledge and understand real-world clinical scenarios and reasoning. Meanwhile, Clinical Camel \citep{clinical-camel} aims to enable LLMs to synthesize plausible clinical records.
\par
\textbf{Electronic Health Records (EHRs)} are digital versions of patients' medical records, offering a more comprehensive overview than clinical notes \citep{hoerbst2010electronic, BUDU2024128253}. EHRs include medical history, diagnoses, medication regimens, immunization dates, allergy information, radiology images, and laboratory test results. This real-time, patient-centric record system provides authorized users with immediate, secure access to patient information. Publicly accessible sources for EHRs include eICU \citep{pollard2018eicu}, MIMIC-IV \citep{johnson2023mimic}, and CPRD \citep{herrett2015data}.
Given the close relation to medical practice and the multidimensional nature of the data, some works \citep{zhongjing, ClinicalGPT} have utilized EHR data to train healthcare domain models, enhancing their usefulness and authenticity. However, different healthcare organizations may use varying EHR systems and record formats, necessitating additional processing to standardize the data. Furthermore, EHRs contain personal medical information that must be anonymized to protect patient privacy.
\par
\textbf{Medical Prescription} is an official directive from a physician or authorized healthcare professional to a pharmacist, detailing the medication to be dispensed to a patient, including the specific drug, dosage, frequency, and administration method \citep{hassan2021medical}. These data reflect real-world prescription practices, providing valuable and authentic information to enhance models' decision-making in healthcare. Some hospitals have integrated electronic medical prescriptions into their EHR systems, significantly facilitating scientific research on this data. Qibo \citep{Qibo} utilizes Traditional Chinese Medicine (TCM) prescriptions as pretraining corpus to incorporate TCM knowledge into medical LLMs. Medical Prescription data is relatively scarce and generally requires collaboration with hospitals to access. Additionally, these datasets may exhibit biases, with certain hospitals or physicians favoring specific drugs or treatments. Errors, omissions, or inaccuracies, such as misspelled drug names or incorrect dosages, can also compromise the effectiveness of model training and application.
\par
\textbf{Doctor-Patient Conversation} data encompasses real-life QA and interactions between doctors and patients, including symptom descriptions, diagnostic information, and patient feedback \citep{maynard2005conversation}. These conversations typically involve multiple rounds of communication, with doctors actively asking questions. Such data can enhance the capabilities of medical LLMs in assisting diagnosis, providing treatment suggestions, asking relevant questions, and understanding context. HuatuoGPT \citep{huatuogpt} uses real medical conversation data in serving instruction fine-tuning to enhance long conversations for medical LLMs. However, accessing this data is challenging due to privacy concerns from hospitals and patients, and the colloquial and emotional nature of these conversations requires future processing.
\par
Overall, the aforementioned professional medical organization corpus sources, generated in real medical scenarios, possess high authenticity and utility. These data provide valuable information for medical LLMs, thereby enhancing their ability to assist in real-world diagnoses. However, these data sources also present several challenges. Primarily, most of the data are proprietary to hospitals and require relevant licenses for access, increasing the difficulty and cost of data acquisition. Additionally, since the data are generated in offline medical settings, they may be incomplete or inaccurate, with potential errors or gaps in medical records. Privacy and security concerns are also significant, as these data may contain sensitive patient information, raising serious legal and ethical issues if leaked or misused.

\subsubsection{Synthetic Data} Synthetic data is generated by prompting an LLM to create content based on its internal knowledge or contextual information \citep{tang2023does}. The common general synthetic datasets are Alpaca-52k \citep{taori2023alpaca}, and ShareGPT \citep{ShareGPT}, which are in QA format. These datasets aim to align customized LLMs with state-of-the-art LLMs at minimal cost. In the medical domain, several LLMs utilize state-of-the-art models to generate training datasets, constructing both single-turn and multi-turn QA formats. Related prompt templates are presented in the GitHub repository.
\par
For single-turn QA data, ChatMed-Consult \citep{ChatGLM-Med} takes medical questions from the internet as input to ChatGPT and then concatenates its outputs to form an instruction fine-tuning dataset. HuatuoGPT \citep{huatuogpt} employs a self-instruct \citep{wang2023self} method, generating a medical instruction dataset by manually creating seed instructions based on specific roles and use cases. Alpacare \citep{alpacare} initially creates a small set of clinician-curated seed tasks, considering four dimensions: topic, view, type, and difficulty level. Each task includes detailed information specific to these dimensions, instructions, and possibly corresponding inputs. GPT-4 then generates new instructions based on these seed tasks, filtering out highly similar ones. Finally, ChatGPT answers each generated instruction to compose the MedInstruct-52k instruction fine-tuning dataset. For multi-turn QA data, HuatuoGPT employs two ChatGPTs: one simulates a doctor and the other as a patient. Real data is utilized to stimulate conversations between these two ChatGPTs, resulting in a multi-turn QA dataset.
\par
The quality of the generated data depends on the effectiveness of the prompts. Well-designed prompts can significantly stimulate the intrinsic knowledge of LLMs, resulting in higher-quality data. However, since this data is not derived from real-world sources, manual checking and correction are typically required to enhance its quality.

\subsection{Data Processing}
After acquiring data from the target source, it may not be immediately suitable for training medical LLMs due to several issues: the presence of errors or irrelevant information, non-standardized data formats, insufficient or overly brief content, language mismatches between the data source and target, tone inconsistencies with the robotic doctor, and lack of professional annotation. Therefore, this survey summarizes the commonly used data processing methods for medical LLMs and classifies them into four categories according to their functions: Data Cleaning, Data Formatting, Data Augmentation, and Translation. Additionally, we summarized the related prompt templates of data processing in the GitHub repository.

\subsubsection{Data Cleaning}
Data cleaning primarily aims to remove noise and correct errors and inconsistencies, improving overall data quality \citep{chu2016data}. We have deeply investigated how existing medical LLM works to clean these data. This process can be categorized by function into Quality Filtering, Sensitive Information Filtering, and De-duplication. 
\par
\textbf{Quality Filtering} enhances data quality by using keywords or special tags to clean the target data. This process can be implemented in various ways, with keyword filtering being a primary method. Keyword filtering helps identify and retain relevant medical domain data, removing non-medical content \citep{wu2020keyword}. By verifying the presence of pertinent medical keywords in a corpus fragment, we can effectively determine its relevance to the desired data. HuatuoGPT-II \citep{Huatuogpt-ii} uses a dictionary-based method to determine if a text fragment belongs to the target domain by evaluating the density of domain-specific words. Similarly, Apollo \citep{Apollo} uses a dictionary-based approach to select medical data from large-scale Chinese, English, and Spanish corpora. MMedLM \citep{MMedLM} employs a rule-based filtering pipeline for extracting target data from multilingual corpora. It constructs a list of 1,200 multilingual medical terms and evaluates text passages based on the count and density of these terms, considering a passage relevant only if both metrics exceed a predefined threshold. Additionally, non-essential sections of medical textbooks, such as the cover, table of contents, and end pages, are excluded to maintain a focus on medical content.
\par
Keyword filtering not only facilitates the extraction of target domain data but also enhances data quality. MedAlpaca \citep{MedAlpaca} ensured data quality by selecting only forum replies that received at least five votes. Qibo \citep{Qibo} improved data quality through character-level and paragraph-level rules to verify the plausibility of individual characters and the semantic continuity of the text. These rules were iteratively updated and refined through sampling and manual checks. InMD-X \citep{InMD-X} initially identified top-ranking journal titles within each internal medicine specialty, based on Journal Citation Reports (JCR), ensuring data quality. Subsequently, it queried these journals via the Pubmed API and extracted papers published since 2010 to capture recent medical knowledge. ChatDoctor \citep{chatdoctor} improved data quality by filtering out short, meaningless conversations. BianQue \citep{Bianque} reduced data noise by removing private content, incomplete JSON data, website links, voice recordings, and auto-responds from medical consulting platform data using regular expressions. Additionally, collected data may include medical advertising with misleading language, potentially biasing medical LLMs. HuatuoGPT-II \citep{Huatuogpt-ii} utilizes ChatGPT to identify advertisements within text and trains a quality classification model to filter out such content from the dataset.
\par
The collected data may contain errors, potentially biasing the model's understanding and leading to inaccuracies in the generated text. This could result in misinformation, such as incorrect disease names, which can confuse users. Additionally, grammatical errors in the collected data may be reflected in the model's output, reducing readability and user experience. To address this, ChatDoctor \citep{chatdoctor} manually filters erroneous content and utilizes LanguageTool for grammatical corrections, while HuatuoGPT \citep{huatuogpt} prompts LLMs to filter data for misspellings, slang, and irrelevant information.
\par
The collected data may include special characters like citation marks or URLs, which lack contextual meaning and are unsuitable for training medical LLMs. PMC-LLaMA \citep{PMC-LLaMA} eliminates irrelevant elements from medical textbooks such as URLs, author lists, redundant information, document content, references, citations, and citation marks for images and tables. MediTron \citep{Meditron} removes URLs, citations, graphics, table separators, and incorrectly formatted characters from clinical guidelines. In processing the literature, author details, bibliographies, acknowledgments, tables, and graphics are excluded, retaining only the main body of each paper. Additionally, it employs ``\#" to denote major sections and ``\#\#" for subsections, replacing citation marks with the title of the reference.
\par
Although the manual review is time-consuming and labor-intensive, it remains the most reliable method for ensuring data quality. OncoGPT \citep{OncoGPT} submits automatically cleaned data for review by domain experts and clinicians to correct inaccuracies and ensure the authority of the answers.
\par
\textbf{Sensitive Information Filtering.} Private information in medical data usually refers to patients' personal identity and health information, which can seriously harm patients' rights if disclosed \citep{ge2021verifiable, lopez2023case}. Moreover, including private attributes like race, gender, and age may introduce bias in certain demographic groups. Common techniques for mitigating privacy concerns include data desensitization, substituting personal identifiers for sensitive information within the text, and data deletion, which involves directly removing sensitive details from the text. However, the latter approach may compromise the integrity and usability of the data. ChiMed-GPT \citep{ChiMed-GPT} and OncoGPT \citep{OncoGPT} remove private information, including personal details, hospital data, and website information, from dialogue data. This process mitigates the impact of data redundancy and non-critical information on model performance.
\par
\textbf{De-duplication} removes duplicates, reduces data redundancy, and enhances the quality and diversity of training data. Additionally, it prevents data leakage between the training and test sets, ensuring a fair evaluation process \citep{borissov2022reducing, fu2021fog}.
HuatuoGPT-II \citep{Huatuogpt-ii} employs a sentence embedding model to transform the corpus into embedding vectors and subsequently eliminates semantically similar texts through an intensive search method. MediTron \citep{Meditron} de-duplicates literature samples by matching paper titles. Apollo \citep{Apollo} performs de-duplication on exam data by removing samples with at least 64 consecutive identical characters to prevent data leakage and ensure accurate assessment results.

\subsubsection{Data Formatting}
The continued pretraining stage requires unstructured data, the instruction fine-tuning stage requires QA pairs (single-turn and multi-turn), and the human alignment stage involves human preference data. The collected data may not be directly suitable for the training stages, necessitating transformation into the appropriate format. We summarize the methods used by medical LLMs to convert data from various structures into three types: Unstructured Data, Instruction Data, and Human Preference Data.
\par
\textbf{Unstructured Data} are sourced from medical encyclopedias, textbooks, academic literature, and guidelines. These data are used in the continued pretraining stage of medical LLMs to enrich domain knowledge and enhance representation capabilities. After a simple truncation process, these unstructured data can effectively support the model's continued pretraining. However, structured data needs to be transformed into unstructured data for continued pretraining. For knowledge graph data, entities, relationships, and attributes are extracted. Subsequently, these information can be transformed into unstructured natural language text using human-designed templates or natural language generation tools like LLM. AntGLM-Med \citep{AntGLM-Med} samples subgraphs at different scales of the knowledge graph and rewrites them as natural language text. QA pairs and NLP task data are commonly stored in JSON format. This format facilitates the extraction of data and labels, enabling the formation of unstructured natural language text using key-value pairs as delimiters. MedicalGPT \citep{MedicalGPT} aggregates questions and answers from the medical encyclopedia QA dataset to create plain text data for continued pretraining, thereby integrating medical knowledge.
\par
\textbf{Instruction Data} consists of question-and-answer(QA) formats, facilitating medical LLMs in comprehending the context and concepts of medical queries. It also aids in familiarizing the models with common expressions and language conventions in the medical domain, resulting in more natural and clinically relevant responses \citep{fleming2024medalign, ouyang2022training}.
Both synthetic data and real doctor-patient conversation data adhere to the question-answer pair format. However, for sources like medical literature, encyclopedias, textbooks, and knowledge graphs, transformation into question-answer pairs requires the application of natural language generation techniques.
MedAlpaca \citep{MedAlpaca} uses ChatGPT to transform paragraph headings from medical textbooks into questions and paragraph content into corresponding answers, thereby generating QA pair dataset. Apollo \citep{Apollo} partitions the corpus data into segments based on basic semantic units, such as chapters in books, paragraphs in website content, and abstracts of papers. Subsequently, it prompts ChatGPT to sequentially generate questions and answers based on these text segments. InMD-X \citep{InMD-X} prompts ChatGPT to extract fundamental medical knowledge from abstracts of medical literature. It subsequently prompts ChatGPT to generate corresponding questions based on these basic medical knowledge. Me-LLaMA \citep{Me-LLaMA} formulates specific prompts for various datasets, including medical literature, clinical notes, clinical guides, Wikipedia, and Knowledge Graphs. It then prompts the LLM to enhance the original questions, thereby creating instruction fine-tuning datasets. Similarly, HuatuoGPT-II \citep{Huatuogpt-ii} prompts ChatGPT to convert the unstructured medical corpus into a question-answer pair format. Additionally, it introduces statistical analysis and semantic recognition techniques to minimize discrepancies between the transformed data and the original source.
Clinical Camel \citep{clinical-camel} presents a dialogue-based knowledge encoding approach, which converts dense medical literature into dialogues and incorporates soft alignment. Initially, it employs a teacher model to convert a segment of literature text into a multi-turn dialogue, with alignment conditions included in the prompts. Subsequently, a student model is tasked with learning the mapping between the literature text and the multi-turn dialogue, resulting in an enhanced student model that improves conversational capabilities.
BiMediX \citep{BiMediX} utilizes samples from a multiple-choice QA dataset and specific prompt words as inputs to ChatGPT, generating a multi-turn QA dataset. DISC-MdeLLM \citep{Disc-medllm} employs LLMs to distill questions and correct answers from the multiple-choice QA dataset. It then integrates these with explanations to produce single-turn QA data.
\par
The above work employs LLMs to perform data formatting, while other works leverage automated methods for the same purpose. ClinicalGPT \citep{ClinicalGPT} utilizes annotations from electronic health records as input for instruction fine-tuning data, with diagnoses as output, forming a dataset of question-answer pairs. GPT-doctor  \citep{GPT-doctor} utilizes package inserts to generate QA pairs covering aspects such as usage, dosage, indications, contraindications, precautions, pharmacological effects, chemical composition, and more. The questions are constructed based on subheadings in the package inserts,  like ``What diseases does [Medicine Name] treat?", and then the answer is the corresponding content of the subheading.
\par
The medical knowledge graph embodies a wealth of structured medical knowledge. Efforts in medical LLMs involve the conversion of knowledge graph data into QA pairs format.
BenTsao \citep{bentsao} first extracts medical knowledge described in natural language from the knowledge graph, and then generates the corresponding instructions and answers based on this knowledge using ChatGPT. DISC-MedLLM \citep{Disc-medllm} initially prompts ChatGPT to convert disease-related knowledge into the $<$instruction, knowledge$>$ format, and then prompts ChatGPT to convert these data into a single-turn medical dialogue scenario $<$user inquiry, physician response$>$. This process aims to augment the diversity and linguistic depth of training datasets. ClinicalGPT \citep{ClinicalGPT} manually designed the corresponding templates to convert each triple from the knowledge graph into data in QA pair format. Likewise, PMC-LLaMA \citep{PMC-LLaMA} formulated QA templates to either describe an entity or predict the relationship between two entities.
\par
\textbf{Human Preference Data} is intended to be used to train the reward model and to guide the LLM in aligning human preferences such as being more doctor-like, patient-friendly, truthful, harmless, and professional \citep{song2024preference, huatuogpt}. This type parallels the QA format, where responses are categorized into chosen and rejected, based on their alignment with the predefined human preferences.
The construction of the human preference dataset is more time-consuming compared to the instruction fine-tuning dataset, Therefore, many studies leverage ChatGPT to facilitate the construction of the human preference dataset.
HuatuoGPT \citep{huatuogpt} utilizes real dialogues to stimulate the generative model into producing diverse responses. These responses undergo evaluation via ChatGPT, considering coherence, consistency with human preferences, and comparison with responses from real doctors. The resultant human preference dataset is structured based on the ratings provided by ChatGPT.
ChiMed-GPT \citep{ChiMed-GPT} and MedicalGPT \citep{MedicalGPT} sample questions from the instruction fine-tuning dataset, then the chosen response is provided by the physician, and the rejected response is provided by the medical LLM BenTsao \citep{bentsao}.
To pursue high-quality human preference data, some of the work uses manual annotation. Zhongjing \citep{zhongjing} employs medical graduate students or clinicians to annotate the human preference dataset with nine dimensions (Accuracy, Safety, Ethics, Comprehension, Clarity, Initiative, Coherence, Consistency, and Warm Tone), yielding a dataset of superior quality. Similarly, ClinicalGPT \citep{ClinicalGPT} employs manual annotation to rank responses to medical inquiries.

\subsubsection{Data Augmentation}
Shorter, colloquial doctor-patient conversation data and multiple-choice QA datasets lacking detailed explanations may compromise the accuracy and professionalism of responses generated by medical LLMs. Consequently, some studies use LLMs to augment the training dataset for addressing these issues.
BianQue \citep{Bianque} prompts ChatGPT to refine doctors' responses over multiple rounds of conversations, enhancing their empathy, professionalism, and coherence. DISC-MedLLM \citep{Disc-medllm} prompts ChatGPT to expand on the original responses, adding detail and logic without altering their meaning. It also eliminates colloquial expressions and revises responses that do not align with the style of the robot doctor, such as advising patients to make appointments. Zhongjing \citep{zhongjing} employs a self-instruct \citep{wang2023self} method to standardize, professionalize, and enhance the friendliness of doctors' responses while preserving the diversity and integrity of the original data. PMC-LLaMA \citep{PMC-LLaMA} prompts GPT-4 to augment medical conversation data by generating synonymous sentences, thereby improving the model's robustness to different instructions. Additionally, it uses ChatGPT to add rationale explanations to multiple-choice QA data. Clinical Camel \citep{clinical-camel} prompts the GPT-4 to retrieve textual resources related to the question, and then supplements the multiple-choice QA data with answer explanations according to those resources. Some studies have used LLMs to augment human preference data. ChiMed-GPT \citep{ChiMed-GPT} enhanced the human preference dataset by incorporating responses from GPT-4 and GPT-3.5. The response format changed from $<$chosen response, rejected response$>$ to $<$chosen response, GPT-4's response, GPT-3.5's response, rejected response$>$, with the degree of alignment to human preferences decreasing in that order.

\subsubsection{Translation}
To make medical LLMs widely accessible across various countries and regions, addressing language differences is crucial. Translating existing medical corpora can overcome this barrier and maximize the utilization of high-quality data. Consequently, some medical LLMs are now focusing on ensuring the accuracy and reliability of translation results. DoctorGLM \citep{doctorglm} prompts ChatGPT to translate the medical corpus data to the target language and then trains a new medical translation model for learning the mapping from the source language to the target language.
BiMediX \citep{BiMediX} employs ChatGPT for English-to-Arabic translation, followed by a quality assessment using ChatGPT to ensure terminology accuracy. If translations do not meet a preset standard, an improvement process begins. This involves providing ChatGPT with the original English text, the current translation, and its evaluation score, prompting it to refine the translation for better consistency with the original. The translation quality is continuously enhanced through iterative feedback. 
Translations that remain unsatisfactory are reviewed and corrected by Arabic-speaking medical experts. Finally, a random sample of high-scoring translations is reviewed to ensure data quality.
To maintain uniformity in language and style within the training corpus, HuatuoGPT-II \citep{Huatuogpt-ii} was inspired by back-translation to standardize the format of the training corpus by prompting ChatGPT. In contrast, the multilingual medical LLM, Apollo \citep{Apollo}, considers cultural differences, beliefs, and taboo phrases across regions. It uses local language medical datasets without translation to avoid potential conflicts.

\section{Training Paradigms} \label{sec:paradigms}
\begin{figure*}[h]
    \centering
    \includegraphics[width=1\linewidth]{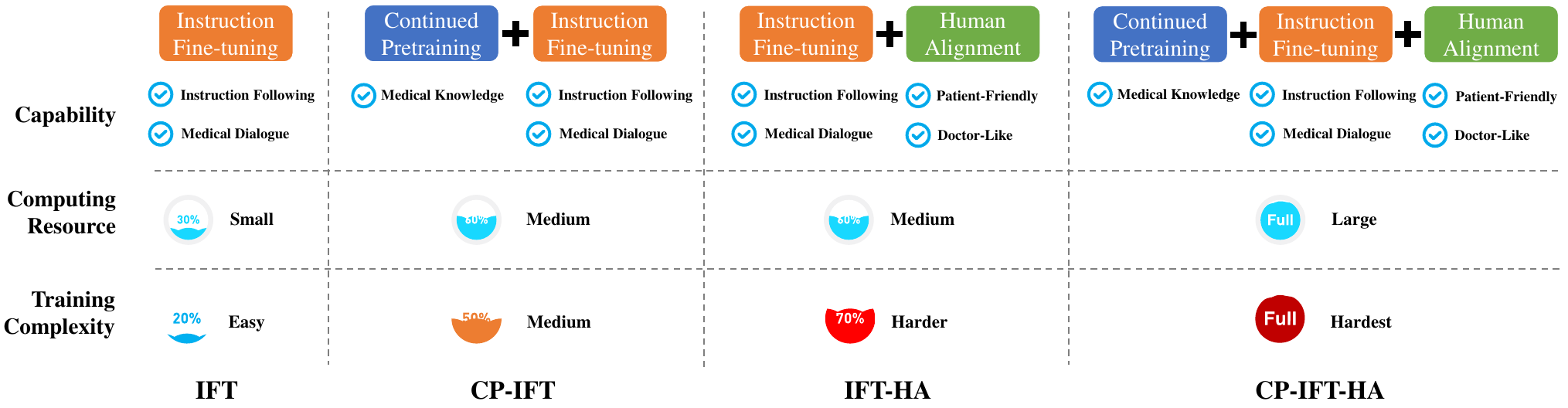}
    \caption{Training Paradigms. The training stage, the achieved capabilities, the required computing resources and the training complexity are provided for each paradigm.}
    \label{fig:method}
\end{figure*}

Transferring from general to medical LLMs involves three stages of continued training: continued pretraining, instruction fine-tuning, and human alignment. However, given that the general LLM has completed training and possesses some medical knowledge alongside its general knowledge, all three stages may not be necessary to train a medical LLM based on the general LLM. Tab. \ref{tab:methods} presents information about the medical LLMs, including the number of parameters, base model, and training stages employed. Medical LLMs learn different capabilities at various training stages and have diverse demands on computational resources and dataset scales. We categorized four paradigms according to the combination of training stages employed by existing medical LLMs, as depicted in Fig. \ref{fig:method}. The characteristics and strategies of each paradigm are described in detail below.

\begin{table*}[htbp]
\fontsize{7}{10}\selectfont
\caption{Detailed Information of Medical Large Language Models. Note: "Para." denotes parameters, "CP" denotes continued pretraining, "IFT" denotes instruction fine-tuning, and "HA" denotes human alignment.}
\label{tab:methods}
\resizebox{\textwidth}{!}{
\begin{tabular}{m{2.5cm}>{\raggedright\arraybackslash}m{3cm}m{1.3cm}m{0.6cm}m{0.6cm}>{\raggedright\arraybackslash}m{2cm}m{0.6cm}m{1.5cm}m{1cm}m{1cm}}
\toprule
Models & Backbone & Para. (B) & CP & IFT & IFT Methods & HA & Preferred Languages & Open Sources & Date \\
\midrule

Med-PaLM \citep{med-palm}               & PaLM \citep{palm}&                540& &                                          $\checkmark$&               Prompt Tuning&                 &                   EN&             & 01/2023\\ \hline 
ChatDoctor \citep{chatdoctor}              &                          LLaMA \citep{llama}&                7& &                                          $\checkmark$ & Full Para.&                 &                   EN&    \href{https://github.com/Kent0n-Li/ChatDoctor}{\textcolor{green}{\checkmark}}     & 03/2023\\ \hline
DoctorGLM  \citep{doctorglm}             &                         ChatGLM \citep{glm130b}&               6&                      &                                         $\checkmark$ &             LoRA&                &                  CN& \href{https://github.com/xionghonglin/DoctorGLM}{\textcolor{green}{\checkmark}}& 04/2023\\  \hline 
BenTsao  \citep{bentsao}      &  LLaMA \citep{llama}&               7&                      &                                         $\checkmark$ &             LoRA&                &                  CN& \href{https://github.com/SCIR-HI/Huatuo-Llama-Med-Chinese}{\textcolor{green}{\checkmark}}& 04/2023\\ \hline 
 ChatGLM-Med \citep{ChatGLM-Med}
& ChatGLM \citep{glm130b}& 6& & $\checkmark$ & LoRA& & CN& \href{https://github.com/SCIR-HI/Med-ChatGLM}{\textcolor{green}{\checkmark}}&04/2023\\ \hline 
 MedAlpaca \citep{MedAlpaca}
& LLaMA \citep{llama}& 7, 13& & $\checkmark$ & Full Para., LoRA& & CN& \href{https://github.com/kbressem/medAlpaca}{\textcolor{green}{\checkmark}}&04/2023\\ \hline 
 PMC-LLaMA \citep{PMC-LLaMA}
& LLaMA2 \citep{llama2}& 13& $\checkmark$ & $\checkmark$ & Full Para.& & CN& \href{https://github.com/chaoyi-wu/PMC-LLaMA}{\textcolor{green}{\checkmark}}& 04/2023\\ \hline 
 HuatuoGPT \citep{huatuogpt}
& Baichuan \citep{baichuan2}, \newline Ziya-LLaMA \citep{ziya-llama}& 7, 13& & $\checkmark$ & Full Para.& $\checkmark$ & CN& \href{https://github.com/FreedomIntelligence/HuatuoGPT}{\textcolor{green}{\checkmark}}&05/2023\\ \hline 
ChatMed-Consult \citep{ChatMed-Consult}& LLaMA \citep{llama}& 7& & $\checkmark$ & LoRA& & CN& \href{https://github.com/michael-wzhu/ChatMed}{\textcolor{green}{\checkmark}}&05/2023\\ \hline 
Med-PaLM 2 \citep{med-palm-2}& PaLM2 \citep{palm2}& -& & $\checkmark$ & -& & EN& &05/2023\\ \hline 
Clinical Camel \citep{clinical-camel} & LLaMA2 \citep{llama2}& 13, 70& & $\checkmark$ & QLoRA& & EN& \href{https://github.com/bowang-lab/clinical-camel}{\textcolor{green}{\checkmark}}&05/2023\\ \hline 
ShenNong-TCM \citep{ShenNong-TCM} & LLaMA \citep{llama}& 7& & $\checkmark$ & LoRA& & CN& \href{https://github.com/michael-wzhu/ShenNong-TCM-LLM}{\textcolor{green}{\checkmark}}&06/2023\\ \hline  
MedicalGPT \citep{MedicalGPT} & Ziya-LLaMA \citep{ziya-llama}, Baichuan-Chat \citep{baichuan2}& 13& $\checkmark$ & $\checkmark$ & LoRA& $\checkmark$ & EN, CN& \href{https://github.com/shibing624/MedicalGPT}{\textcolor{green}{\checkmark}}&06/2023\\ \hline 
ClinicalGPT \citep{ClinicalGPT} & BLOOM \citep{bloom}& 7& & $\checkmark$ & LoRA& $\checkmark$ & CN& &06/2023\\ \hline 
DISC-MedLLM \citep{Disc-medllm}
& Baichuan \citep{baichuan2}& 13& & $\checkmark$ & Full Para.& & CN& \href{https://github.com/FudanDISC/DISC-MedLLM}{\textcolor{green}{\checkmark}}& 08/2023\\ \hline 
     Zhongjing \citep{zhongjing}
& Ziya-LLaMA \citep{ziya-llama}& 13& $\checkmark$ & $\checkmark$ & LoRA& $\checkmark$ & CN& \href{https://github.com/SupritYoung/Zhongjing}{\textcolor{green}{\checkmark}}&08/2023\\ \hline 
      BianQue \citep{Bianque}
& ChatGLM \citep{glm130b}& 6& & $\checkmark$ & Full Para.& & CN& \href{https://github.com/scutcyr/BianQue}{\textcolor{green}{\checkmark}}&10/2023\\ \hline 
       Alpacare \citep{alpacare}
& LLaMA \citep{llama}& 7, 13& & $\checkmark$ & Full Para.& & EN& \href{https://github.com/scutcyr/BianQue}{\textcolor{green}{\checkmark}}&10/2023\\ \hline 
        Qilin-Med \citep{Qilin-med}
&Baichuan \citep{baichuan2} & 7 & $\checkmark$ & $\checkmark$ & LoRA& $\checkmark$ & CN & \href{https://github.com/williamliujl/Qilin-Med}{\textcolor{green}{\checkmark}}&10/2023\\ \hline 
 Taiyi \citep{Taiyi}
& Qwen \citep{qwen} & 7 & & $\checkmark$ & QLoRA& &EN, CN & \href{https://github.com/DUTIR-BioNLP/Taiyi-LLM}{\textcolor{green}{\checkmark}}&11/2023\\ \hline 
 ChiMed-GPT \citep{ChiMed-GPT}
& Ziya-LLaMA \citep{ziya-llama}& 13 & $\checkmark$ & $\checkmark$ & Full Para.& $\checkmark$ & CN& \href{https://github.com/synlp/ChiMed-GPT}{\textcolor{green}{\checkmark}}&11/2023\\ \hline 
 MediTron \citep{Meditron}
& LLaMA2 \citep{llama2} & 7, 70 & $\checkmark$ & $\checkmark$ & Full Para.& & EN& \href{https://github.com/epfLLM/meditron}{\textcolor{green}{\checkmark}}&11/2023\\ \hline 
 HuatuoGPT-II \citep{Huatuogpt-ii}
&Baichuan2 \citep{baichuan2}, \newline Yi \citep{yi} & 7, 13, 34 & & $\checkmark$ & Full Para.& & CN & \href{https://github.com/FreedomIntelligence/HuatuoGPT-II}{\textcolor{green}{\checkmark}}&12/2023\\ \hline 
 AntGLM-Med \citep{AntGLM-Med}&GLM \citep{glm} &10 & $\checkmark$ & $\checkmark$ & Full Para., LoRA, Cpoly& &EN, CN &  &12/2023\\ \hline 
GPT-doctor \citep{GPT-doctor} & Baichuan2-Chat \citep{baichuan2} &13 & & $\checkmark$ & LoRA& & CN & &12/2023\\ \hline 
EpilepsyLLM \citep{EpilepsyLLM} & LLM-JP \citep{llm-jp}, LLaMA \citep{llama} & 1.3, 7 & & $\checkmark$ & LoRA& & EN, JP & \href{https://github.com/masa3141/japanese-alpaca-lora}{\textcolor{green}{\checkmark}}& 01/2024 \\ \hline 
 BioMistral \citep{BioMistral}
&Mistral-Instruct \citep{mistral} & 7 & $\checkmark$ & $\checkmark$ & QLoRA& & Multilingual & \href{https://huggingface.co/BioMistra}{\textcolor{green}{\checkmark}}&02/2024 \\ \hline
MMedLM \citep{MMedLM}& InternLM \citep{internlm} & 7 & $\checkmark$ & $\checkmark$ & Full Para., LoRA& & Multilingual& \href{https://github.com/MAGIC-AI4Med/MMedLM}{\textcolor{green}{\checkmark}}&02/2024\\ \hline
InMD-X \citep{InMD-X}& Neural-Chat \citep{neural-chat} & 7& $\checkmark$ & $\checkmark$ & Full Para., LoRA& &EN & &02/2024\\ \hline
Me-LLaMA \citep{Me-LLaMA} &LLaMA2 \citep{llama2} & 13, 70& $\checkmark$ & $\checkmark$ & LoRA& &EN &\href{https://github.com/BIDS-Xu-Lab/Me-LLaMA}{\textcolor{green}{\checkmark}} &02/2024\\ \hline
JMLR \citep{JMLR} &LLaMA2 \citep{llama2} &7, 13 & $\checkmark$ & $\checkmark$ & -& &EN & &02/2024 \\  \hline
BiMediX \citep{BiMediX} &Mixtral-8x7B \citep{mixtral8-7B} & 8x7 & &$\checkmark$ &QLoRA & &EN, Arabic &\href{https://github.com/mbzuai-oryx/BiMediX}{\textcolor{green}{\checkmark}} &02/2024\\  \hline
OncoGPT \citep{OncoGPT} &LLaMA \citep{llama} & 7 & &$\checkmark$ &LoRA & &EN &\href{https://github.com/OncoGPT1/OncoGPT1}{\textcolor{green}{\checkmark}} &02/2024\\  \hline
Apollo \citep{Apollo} &Qwen \citep{qwen}, \newline Gemma \citep{gemma}, \newline Yi \citep{yi}    & 0.5, 1.8, 2, 6, 7 & $\checkmark$&$\checkmark$ &Full Para. & &Multilingual &\href{https://github.com/FreedomIntelligence/Apollo}{\textcolor{green}{\checkmark}}  &03/2024 \\ \hline
Qibo \citep{Qibo} & LLaMA \citep{llama} & 7, 13 &$\checkmark$ &$\checkmark$ &Full Para. & &CN & &03/2024\\ \hline
Hippocrates \citep{hippocrates} & LLaMA2 \citep{llama2},\newline Mistral \citep{mistral} & 7 &$\checkmark$ & $\checkmark$  & LoRA & $\checkmark$ & EN, CN & \href{https://cyberiada.github.io/Hippocrates}{\textcolor{green}{\checkmark}} & 04/2024 \\ \hline
MING-MOE \citep{MING-MOE} & Qwen1.5-Chat \citep{qwen} & 1.8, 4, 7, 14 & & $\checkmark$  & LoRA & & EN, CN & \href{https://github.com/MediaBrain-SJTU/MING}{\textcolor{green}{\checkmark}} & 04/2024 \\ \hline
Lingdan \citep{lingdan} & Baichuan2 \citep{baichuan2} & 13 & $\checkmark$ & $\checkmark$ & QLoRA & &EN, CN &  \href{https://github.com/TCMAI-BJTU/LingdanLLM}{\textcolor{green}{\checkmark}} & 04/2024 \\ \hline
Aloe \citep{aloe} & Mistral \citep{mistral}, \newline LLaMA3 \citep{llama3} & 7, 8& & $\checkmark$ &Full Para. &$\checkmark$  & EN & \href{https://huggingface.co/HPAI-BSC/Llama3-Aloe-8B-Alpha}{\textcolor{green}{\checkmark}} & 05/2024 \\ \hline
PediatricsGPT \citep{PediatricsGPT} & Baichuan2 \citep{baichuan2} & 7,13 & $\checkmark$ & $\checkmark$ & Full Para., LoRA & $\checkmark$ & CN & & 06/2024 \\ \hline
Aqulia-Med \citep{Aqulia-Med} & Aquila \citep{Aquila} & 7 & $\checkmark$ & $\checkmark$ & Full Para. & $\checkmark$ & EN, CN & \href{https://huggingface.co/BAAI/AquilaMed-RL}{\textcolor{green}{\checkmark}} & 06/2024 \\
\bottomrule
\end{tabular}
}
\end{table*}

\subsection{IFT Paradigm}
Instruction fine-tuning  (IFT), also termed supervised fine-tuning  (SFT), refers to the parameter fine-tuning of pretrained LLMs using instruction data (e.g., QA pairs). During the training process, only the loss of response content is considered. After instruction fine-tuning, medical LLMs can demonstrate strong medical instruction following and zero-shot learning ability. It can effectively grasp instructions from medical practitioners or patients, thereby enhancing doctors' clinical efficiency and addressing patients' needs efficiently.
\par
Medical LLMs depicted in Tab. \ref{tab:methods} all employ instruction fine-tuning to enhance their capacity for addressing medical issues. This paradigm leverages the extensive knowledge base inherent in general LLMs, requiring only instruction fine-tuning to adapt LLMs to specific medical tasks. Additionally, the emergence of Parameter Efficient Fine-Tuning (PEFT) \citep{lialin2023scaling} accelerates the training process of instruction fine-tuning and reduces the demand for computational resources.
\textbf{Strengths}: Instruction fine-tuning can improve large medical language models' performance in instruction following adherence with minimal computational resources and time consumption. \textbf{Weaknesses}: Without continued pretraining and human alignment, the knowledge understanding of the model may be limited, and its response may not accurately simulate the characteristics of the physician.

\subsubsection{Parameter Efficient Fine-Tuning} 
As can be seen from Tab. \ref{tab:methods}, there are two types of instruction fine-tuning methods used in medical LLMs: one is full parameter fine-tuning, and the other is PEFT \citep{lialin2023scaling}. Full parameter fine-tuning involves training and updating all parameters related to the medical LLM during instruction fine-tuning. Conversely, PEFT reduces the number of model parameters requiring training, thereby reducing the performance demands of the necessary equipment for training while ensuring that the fine-tuned model's performance is comparable to that of full parameter fine-tuning. The most commonly used parameter-efficient fine-tuning method in the medical LLMs is Low-Rank Adaptation (LoRA) \citep{hu2021lora}. LoRA approximates parameter updates for each layer by adding low-rank decomposition matrices while freezing the original weights, thereby reducing the number of training parameters and the training time. Another commonly used method is QLoRA \citep{dettmers2024qlora}, which is a parameter quantization technique based on the LoRA method. In QLoRA, the original model parameters are quantized to 4-bit, while the newly added low-rank decomposition matrix parameters are saved as 16-bit for training. This approach further reduces the performance requirements of the training equipment and lowers the threshold for developing medical LLMs in medical institutions.
In addition, there are many PEFT methods, such as Adapter Tuning \citep{houlsby2019parameter}, Prefix Tuning \citep{li2021prefix}, Cpoly \citep{wang2023customizable}, Prompt Tuning \citep{lester2021power} and, P-tuning \citep{liu2023gpt,liu2022p}, which can also reduce the performance requirements and time consumption of the training equipment. However, these methods are used less frequently than LoRA.

\subsubsection{Two-stage Fine-tuning Strategy}
Given the diverse sources of data reflecting various aspects of medical knowledge, a one-stage instruction fine-tuning approach may not fully exploit these distinct data sources.
Consequently, some efforts in developing medical LLMs have begun to adopt a two-stage instruction fine-tuning approach.
DISC-MedLLM \citep{Disc-medllm} adopted a two-stage training strategy for the instruction fine-tuning stage. The first stage involved training on large-scale instruction data, utilizing various datasets to inject medical domain knowledge and dialogue capabilities into the model. The second stage focused on a small amount of high-quality data, generated with the assistance of ChatGPT and filtered and guided by humans, characterized by matching human behavioral preferences. This stage aims to align the behavior of the medical LLMs with human preferences.
Taiyi \citep{Taiyi} considers that the instruction dataset comprises multiple tasks, and one-stage instruction fine-tuning can lead to interference between tasks, necessitating a two-stage strategy. The first stage uses conventional NLP task datasets, such as information extraction and text classification. The second stage uses QA and dialogue task datasets, training these alongside the first stage data. This two-stage training allows the model to initially focus on processing non-generative tasks and then, in the second stage, enhance its cross-task and dialogue task processing capabilities.
OncoGPT \citep{OncoGPT} is first trained on Alpaca and doctor-patient communication data from online medical consultation platforms for instruction fine-tuning to gain basic medical conversational skills.  It is then further trained on real doctor-patient oncology conversation data thoroughly proofread and reviewed by experts and clinicians to improve cancer information processing and conversation skills.

\subsection{CP-IFT Paradigm}
Continued pretraining, also known as continued domain-adaptive pretraining, involves further pretraining of LLMs on domain-specific corpora to enrich domain knowledge \citep{ke2023continual}. The CP-IFT paradigm is that the general LLM initially performs continued pretraining using unstructured medical data to enhance its medical knowledge, adapt to the specific terminology, language style, and text structure of the medical domain, and gain a deeper understanding of medical expertise and complex concepts. It then performs instruction fine-tuning on medical instruction data to improve its ability to follow medical instructions. A few studies follow the CP-IFT paradigm to train medical LLMs on top of general LLMs. Additionally, there are several studies \citep{Meditron, BioMistral, MMedLM} where their instruction fine-tuning is executed on the training set of the evaluation benchmarks for overcoming forgetfulness of instruction following ability.
\par
The \textbf{advantage} of the CP-IFT paradigm is that, compared to the IFT paradigm, the model learns more medical knowledge and patterns during the continued pretraining, enhancing its generalization capabilities and enabling faster convergence during the instruction fine-tuning for various medical tasks and sub-domains. However, the \textbf{disadvantage} of the CP-IFT paradigm is that it requires more training data, extensive computational resources, and longer training times compared to the IFT paradigm. Additionally, if the data used for continued pretraining is irrelevant or significantly different from the instruction fine-tuning tasks, the model may learn irrelevant information, which can degrade performance on downstream tasks.

\subsubsection{Model Merging}
Model Merging refers to the merging of two or more models in order to improve the generalization performance of the model without additional training. Most methods focus on fusing model parameters. BioMistral \citep{BioMistral} merges the pretrained medical LLMs with the base model using three methods (TIES \citep{yadav2023tiesmerging}, DARE \citep{yu2024language}, and SLERP \citep{shoemake1985animating}) for weight merging respectively, to improve the general domain capabilities of the medical LLMs. Apollo \citep{Apollo} uses Proxy Tuning \citep{liu2024tuning} to indirectly guide the adjustment of general LLMs with a large number of parameters by utilizing the logits (outputs) of the pre-fine-tuning model (without medical knowledge injected) and the fine-tuned model (with medical knowledge injected). Experimental findings indicate that proxy tuning enhances the model's performance on multiple language benchmarks beyond the English benchmark.

\subsubsection{Combining Pretraining Data with Instruction Data}
Transforming pretraining data into instruction data and jointly training it with existing instruction data can enhance the effectiveness and stability of the training process. \citep{he2023survey}. 
Building upon this concept, HuatuoGPT-II \citep{Huatuogpt-ii} transforms its pretrained medical data into an instruction format, creating a pre-trained instruction dataset. This dataset is then jointly trained with existing instruction data. Additionally, it proposes a priority sampling strategy to mitigate the challenges associated with mixing data sources during LLM training. Similarly, Apollo \citep{Apollo} rewrites the pretraining corpus into the data format of QA and adopts the same data mixing and sampling techniques as HuatuoGPT-II. This approach facilitates a smoother transition from continued pretraining to instruction fine-tuning.

\subsection{IFT-HA Paradigm}
The IFT-HA paradigm adopts a two-stage approach. In the first stage, instruction fine-tuning is employed to augment the LLM's ability to follow medical instructions and engage in medical dialogue. Subsequently, the second stage focuses on human alignment, ensuring the LLM's responses adhere to established medical alignment principles. 
In the general domain, human alignment aims to ensure that LLMs operate in accordance with human values, genuine intentions, and established social ethics. The primary alignment principles are helpful, honest, and harmless. Within the medical domain, these core principles are further extended to encompass patient-friendly and doctor-like (professional and interactive diagnostic) \citep{huatuogpt}. 
\par
To achieve human alignment in the medical domain, a prevalent approach is Reinforcement Learning from Human Feedback (RLHF) \citep{ouyang2022training}. This methodology entails a two-stage process. Initially, a reward model is constructed on top of an instruction fine-tuned medical LLM, utilizing manually labeled human preference data. Subsequently, this reward model is combined with either PPO \citep{schulman2017proximal} or the rejection sampling strategy \citep{llama2} to perform reinforcement learning training, thereby aligning medical LLMs with predetermined principles. Considering the difficulty of collecting human preference data, HuatuoGPT, motivated by RLHF and RLAIF \citep{lee2023rlaif}, proposes Reinforcement Learning with Mixed Feedback (RLMF). RLMF leverages ChatGPT to generate a portion of the human preference data, effectively merging the strengths of ChatGPT and real doctors. Consequently, medical LLMs are guided to align responses with both ChatGPT and medical expertise. Qilin-Med \citep{Qilin-med} adopts Direct Preference Optimization (DPO) \citep{rafailov2024direct}, a non-reinforcement learning approach, to accomplish human alignment training for medical LLMs. Unlike reinforcement learning methods that rely on reward modeling, DPO directly establishes the connection between the decision function and the reward function, obviating the need for explicit reward modeling.
\par
Currently, human alignment training is built on models that have undergone instruction fine-tuning. The IFT-HA paradigm offers several \textbf{advantages}: medical LLMs exhibit enhanced instruction following capabilities, yielding more accurate responses to user queries and physician instructions. Additionally, IFT-HA fosters alignment with human values, promoting the provision of accurate and reliable medical information, clear explanations, and the avoidance of harmful suggestions while safeguarding user privacy. However, the IFT-HA paradigm presents \textbf{disadvantages}. The human alignment training process is intricate, susceptible to instability, and highly sensitive to hyperparameter selection.  Furthermore, RLHF is highly dependent on the performance of the reward model, which in turn is dependent on the quality of the expensive human preference data. Consequently, developers opting for this paradigm need substantial engineering experience.

\subsection{CP-IFT-HA Paradigm}
The CP-IFT-HA paradigm injects medical knowledge at different dimensions into the general LLM in three stages: continued pretraining, instruction fine-tuning, and human alignment. Due to the complexity and substantial computational resources required, few medical LLM efforts have adopted this paradigm. However, medical LLMs trained using this paradigm demonstrate superior medical knowledge comprehension, instruction following, and alignment with human preferences compared to other paradigms \cite{zhongjing}.
\par
Compared to the CP-IFT paradigm, the paradigm introduces the human alignment stage. This enhancement empowers medical LLMs to generate outputs that are more aligned with medical principles and human expectations, thereby reducing potentially harmful outputs and fostering greater caution when answering medical questions. In particular, the paradigm can better protect patient privacy when dealing with sensitive medical information. 
Compared to the IFT-HA paradigm, CP-IFT-HA paradigm incorporates additional stages of continued pretraining, resulting in medical LLMs equipped with a more robust foundation in medical knowledge. The enhanced knowledge enables models to possess greater adaptability within the medical domain. Consequently, models using this paradigm are able to provide more accurate and insightful answers, which is especially important in medical domains that involve highly specialized and precise questions.

\subsection{Summary}
The choice of paradigms should be based on the demands, resources, and time constraints of a specific project. In the medical field, where high expertise and accuracy are paramount, the CP-IFT-HA paradigm is more advantageous. However, the IFT paradigm can better meet practical needs in scenarios requiring rapid deployment with limited resources. The CP-IFT and IFT-HA paradigms strike a balance between the performance of medical LLMs and the complexity of development. Specifically, the CP-IFT paradigm focuses on the comprehension and processing of medical knowledge, ensuring that the model can accurately grasp and apply relevant information. While the IFT-HA paradigm emphasizes the safety and doctor-like professionalism of responses, ensuring patient safety and trust.

\section{Evaluations} \label{sec:evaluations}
Given the specialized and sensitive nature of medical information, erroneous or unsafe responses could potentially mislead and harm patients. Hence, it is crucial to accurately and comprehensively evaluate the performance of medical LLMs \citep{chang2023survey, rydzewski2024comparative}. We systematically summarize the existing evaluation methods for medical LLMs and classify them into two main perspectives: Machine Evaluation and Human-Centric Evaluation. The former focuses on using deterministic evaluation metrics to measure the performance of medical LLMs, while the latter evaluates them from a human perspective. Tab. \ref{tab:eva} presents the evaluation protocols and metrics for existing medical LLMs.

\begin{table*}[H]
\fontsize{7}{9}\selectfont
\caption{Evaluation Setting Details for the Medical Large Language Model. The abbreviations are as follows: STQA for Single-turn QA, MTQA for Multiple-turn QA, MCQA for Multiple-choice QA, NLP for Conventional Natural Language Processing Tasks, Zero. for Zero-shot Learning, Few. for Few-shot Learning, and SFT for Task-specific Supervised Fine-tuning.}
\label{tab:eva}
\begin{tabular}{m{2.5cm}>{\raggedright\arraybackslash}m{2cm}>{\raggedright\arraybackslash}m{2.5cm}>{\raggedright\arraybackslash}m{1.6cm}m{1cm}>{\raggedright\arraybackslash}m{2.6cm}m{1.5cm}}
\toprule
\multirow{2}{*}{Models} & \multirow{2}{*}{Types} & \multicolumn{2}{c}{Machine Evaluation}  & \multicolumn{3}{c}{Human-Centric Evaluation} \\ 
\cmidrule(r){3-4}\cmidrule(r){5-7}
                        &  & Metrics& Protocols & Evaluator & Dimensions  & Protocols 	\\
\midrule

Med-PaLM \citep{med-palm}               & STQA, MCQA &                Accuracy& Zero., Few.&                                          Human&               Professional, Safe, Helpful&                 Individual\\ \hline 
ChatDoctor \citep{chatdoctor}              &                          STQA&                BERTScore& Zero. &                                          Human& &                  Case Study\\ \hline
DoctorGLM  \citep{doctorglm}             &                         &               &                      &                                         Human&             &                Case Study\\  \hline 
BenTsao  \citep{bentsao}      &  STQA&               &                      &                                         Human&             Safe, Fluent, Helpful&                Individual\\ \hline 
 ChatGLM-Med \citep{ChatGLM-Med}
& & & & Human& &Case Study\\ \hline 
 MedAlpaca \citep{MedAlpaca}
& MCQA& Accuracy& Zero. & & & \\ \hline 
 PMC-LLaMA \citep{PMC-LLaMA}
& MCQA& Accuracy& Zero., SFT& & &  \\ \hline 
 HuatuoGPT \citep{huatuogpt}
& STQA, MTQA& BLEU, ROUGE, GLEU, Distinct& Zero.& Human, LLM& Professional, Proactive, Doctor-like, Patient-friendly& Pairwise\\ \hline 
ChatMed-Consult \citep{ChatMed-Consult}& & & & Human& & Case Study\\ \hline 
Med-PaLM 2 \citep{med-palm-2}& STQA, MCQA& Accuracy& Zero., Few.& Human& Professional, Safe, Helpful& Individual,  Pairwise\\ \hline 
Clinical Camel \citep{clinical-camel} & MCQA& Accuracy& Zero., Few.& & & \\ \hline 
ShenNong-TCM \citep{ShenNong-TCM} & & & & Human& & Case Study\\ \hline  
MedicalGPT \citep{MedicalGPT} & & & & Human& &Case Study\\ \hline 
ClinicalGPT \citep{ClinicalGPT} & STQA, MCQA& BLEU, ROUGE,  GLEU, Accuracy& Zero.& LLM& Professional, Safe, Helpful& Pairwise\\ \hline 
DISC-MedLLM \citep{Disc-medllm}
& MTQA, MCQA& Accuracy& Zero., Few.& LLM& Proactive, Professional, Helpful&  Individual\\ \hline 
 Zhongjing \citep{zhongjing}
& STQA, MTQA& & & Human, LLM& Professional, Safe, Fluent& Pairwise, Case Study\\ \hline 
  BianQue \citep{Bianque}
& MTQA& BLEU, ROUGE, PQA& Zero.& Human& & Case Study\\ \hline 
   Alpacare \citep{alpacare}
& STQA, MTQA& Accuracy, ROUGE& Zero., Few.& Human, LLM& Professional, Helpful& Pairwise, Case Study\\ \hline 
    Qilin-Med \citep{Qilin-med}
& STQA, MTQA& F1-Score, BLEU, ROUGE& Zero., Few., SFT& Human& & Case Study\\ \hline 
Taiyi \citep{Taiyi}
& MTQA, NLP& F1-Score, Accuracy& Zero.& Human& &Case Study\\ \hline 
ChiMed-GPT \citep{ChiMed-GPT}
& STQA, MTQA, MCQA, NLP& F1-Scores,
Accuracy, BLEU, ROUGE& Zero., Few.& Human& & Case Study\\ \hline 
MediTron \citep{Meditron}
& MCQA& Accuracy& Zero., Few., SFT& & & \\ \hline 
HuatuoGPT-II \citep{Huatuogpt-ii} & STQA, MTQA, MCQA& Accuracy& Zero.& Human, LLM& Professional, Helpful, Patient-friendly, Fluent& Pairwise, Case Study\\ \hline 
AntGLM-Med \citep{AntGLM-Med}& MCQA& Accuracy& Few., SFT& & & \\ \hline 
GPT-doctor \citep{GPT-doctor} & STQA& & & Human, LLM& Professional, Fluent, Patient-friendly, Doctor-like, & Individual, Pairwise\\ \hline 
EpilepsyLLM \citep{EpilepsyLLM} & STQA & BLEU,  ROUGE, METEOR, SPICE & Zero. & & & \\ \hline
BioMistral \citep{BioMistral} & MCQA& Accuracy& Zero., Few., SFT& & & \\ \hline
MMedLM \citep{MMedLM}& STQA, MCQA& Accuracy, BLEU, ROUGE, BERTScore& Zero., SFT& Human, LLM& Professional& Pairwise\\ \hline
InMD-X \citep{InMD-X}& & & & Human& & Case Study\\ \hline
Me-LLaMA \citep{Me-LLaMA} & MCQA, NLP& Accuracy, F1-score, ROUGE, BERTScore& Zero., Few., SFT& & & \\ \hline
JMLR \citep{JMLR} & & & & Human& &  Case Study\\  \hline
BiMediX \citep{BiMediX} & MCQA& Accuracy& Zero.& & & \\  \hline
OncoGPT \citep{OncoGPT} & STQA& BERTScore& Zero.& Human& & Case Study\\  \hline
Apollo \citep{Apollo} & MCQA& Accuracy& Few.& & & \\ \hline
Qibo \citep{Qibo} & STQA, MCQA, NLP & Accuracy, ROUGE& Zero.& Human, LLM& Professional, Safe, Fluent& Pairwise\\ \hline
Hippocrates \citep{hippocrates} & MCQA & Accuracy & Zero., Few. & & & \\  \hline
MING-MOE \citep{MING-MOE} & STQA, MCQA & F1, ROUGE, Accuracy & Zero.& Human & & Case Study\\ \hline
Lingdan \citep{lingdan} & STQA, MTQA & Top@K & Zero., & & & \\ \hline
Aloe \citep{aloe} & STQA, MCQA & Accuracy, ASR &  Zero., Few. & & & \\ \hline
PediatricsGPT \citep{PediatricsGPT} & STQA & BLEU, ROUGE, GLEU, Distinct & Zero. & Human, LLM & Professional, Helpful, Safe, Fluent & Pairwise, Case Study \\ \hline
Aqulia-Med \citep{Aqulia-Med} & STQA, MTQA, MCQA & Accuracy & Zero., Few. & LLM & Professional, Fluent, Helpful & Pairwise \\
\bottomrule
\end{tabular}
\end{table*}

\subsection{Machine Evaluation}
\subsubsection{Benchmark Types}
Machine Evaluation assesses the performance of medical LLMs in handling natural language tasks, utilizing established benchmarks for natural language understanding and generation. These benchmarks provide explicit, static evaluation criteria for quantitatively measuring the performance of medical LLMs.
\par
As shown in Tab. \ref{tab:eva}, few studies have used natural language understanding tasks as evaluation benchmarks, such as Medical Named Entity Recognition, Medical Relationship Extraction, and Medical Text Classification. General LLMs typically include natural language understanding tasks in training sets, providing them with capability in this area. This evaluation aims to assess whether Medical LLMs retain natural language understanding ability after adaptation to medical information.
Me-LLaMA \citep{Me-LLaMA} and Taiyi \citep{Taiyi} utilize natural language understanding benchmarks to evaluate the performance of the proposed medical LLM. They first perform task-specific supervised fine-tuning on the training sets of these benchmarks, followed by performance evaluation on the test sets. Common evaluation metrics for these tasks include Accuracy, F1-Score \citep{powers2020evaluation}, and BERTScore \citep{zhang2019bertscore}. Additionally, a limited number of studies employed the evaluation metrics METEOR \citep{banerjee2005meteor}, SPICE \citep{anderson2016spice}, and Attack Success Rate (ASR) \cite{aloe}.
\par
Many studies have used QA tasks (single-turn, multi-turn, and multiple-choice) in natural language generation as evaluation benchmarks to measure the quality of response content and the performance of processing medical QA tasks for medical LLMs. In particular, the \textbf{single-turn QA} benchmark involves a single question and a single answer without continuous dialogue. It allows direct assessment of the model's performance and knowledge coverage on specific medical questions. And it is applicable to a wide range of medical topics, from simple definitions to complex diagnostic problems. The \textbf{multi-turn QA} benchmark involves consecutive questions and answers, such as a dialogue between a doctor and a patient. This benchmark assesses the ability of medical LLMs to perform medical diagnosis, understand context (ensuring answers are based on previous dialogue), and engage in interactive QA (testing the model's ability to interact in real-world medical scenarios and maintain fluent dialogue). The \textbf{multiple-choice QA} benchmark presents a question and multiple options, requiring the model to identify the correct answer. With explicit correct answers, this benchmark allows precise quantification of the medical LLM's capability. Additionally, some works not only evaluate the options outputted by the model but also assess reasoning content. The varied multiple-choice questions evaluate the model's depth of knowledge across diverse medical domains and its logical reasoning ability. The commonly used evaluation metrics for natural language generation tasks are Accuracy, BLEU \citep{papineni2002bleu}, ROUGE \citep{lin2004rouge}, Distinct \citep{li2016diversity} and GLEU \citep{wu2016google}. In addition to traditional evaluation metrics, certain studies have introduced evaluation metrics tailored to healthcare dialogue contexts. For instance, BianQue \citep{Bianque} proposed a novel evaluation metric named Proactive Questioning Ability (PAQ) to assess the capability of LLMs in proactively asking and querying questions to users.
\par
The majority of medical QA benchmarks comprise existing medical QA datasets, with only a limited number of benchmarks specifically gathered for particular evaluations. Note that the vast datasets employed in training LLMs may overlap between training and test sets. Therefore, data de-duplication and data leakage checks are necessary to ensure the fairness of the test results.

\subsubsection{Protocols}
Medical LLMs are commonly assessed through three protocols: Zero-shot, Few-shot, and Task-specific Supervised Fine-tuning, aiming to investigate model performance across varying levels of contextual information.
\par
\textbf{Zero-shot} protocol is to evaluate the ability of a model to directly perform a medical task with an unknown distribution while assessing its generalization ability and adaptability. It only requires a natural language description of the downstream task, without providing specific examples, to ask the model to understand the medical task and use the intrinsic knowledge to accomplish this task. In the natural language understanding task, Me-LLaMA \citep{Me-LLaMA} and Taiyi \citep{Taiyi} designed prompt templates for each downstream task to guide the output format of the model to meet the expected requirements. For the prompt of QA tasks, it usually gives a role to the LLM and then guide the model's output style in line with expectations. For instance, a prompt might be, "You are an experienced doctor, and please respond to the patient's inquiries with patience and kindness. \#\#\# Patient: $<inputs>$ \#\#\# Doctor:". Moreover, addressing medical issues often necessitates intricate multi-step reasoning. Hence, certain studies have employed prompt strategies to enhance model performance. For instance, MediTron \citep{Meditron} utilizes the prompt strategy Zero-shot Chain-of-Thought \citep{kojima2022large} to guide the model towards a more comprehensive understanding of medical problems, facilitating the generation of reasoning steps and answers. AntGLM-Med \citep{AntGLM-Med} introduces a simple prompt strategy called Verification-of-Choice (VoC), tailored for the multiple-choice QA benchmark. VoC operates under the assumption that each choice is correct, prompting the LLM to provide the corresponding explanation. Subsequently, these explanations serve as a context for the LLM to identify the inconsistencies and give the final answer.
\par
\textbf{Few-shot} protocol evaluates the performance of LLMs on downstream tasks involving a limited number of examples. Typically, examples are derived from the training set in the downstream task, including both inputs and corresponding outputs. The medical LLM then utilizes these examples, in conjunction with its intrinsic medical knowledge, to generate appropriate outputs in accordance with the provided instructions. 
Beyond simply employing a limited number of examples as context, several studies \citep{Meditron} have incorporated prompting strategies like Chain-of-Thought \citep{wei2022chain} and Self-Consistency Chain-of-Thought (SC-CoT) \citep{wang2022self} to stimulate the reasoning capabilities of medical LLMs. These strategies aim to enhance the accuracy of outputs and the rationality of reasoning processes.  Additionally, prompting strategies can facilitate the identification and correction of potential errors and knowledge gaps in medical LLMs, particularly in complex medical tasks.
\par
\textbf{Task-specific supervised fine-tuning} involves training a medical LLM on a benchmark training set and evaluating it on a corresponding test set \citep{BioMistral, Qilin-med}. By training the medical LLM on these specific medical tasks, such as drug named entity recognition and medical literature comprehension, the generalization performance and knowledge transfer ability of the model can be evaluated. Two types of training approaches, parameter-efficient fine-tuning, and full-parameter fine-tuning, are used to evaluate the performance of the model under two different levels of medical knowledge injections \citep{MMedLM}.

\subsection{Human-Centric Evaluation}
Given the inherent diversity and complexity of model responses, the machine evaluation may not adequately captures the performance of medical LLMs, particularly in assessing the usefulness and safety of outputs. To address this limitation, researchers have increasingly adopted Human-Centric Evaluation approaches. It provides a more realistic assessment of medical LLMs in real-world applications, reflecting human judgment and expectations. 

\subsubsection{Evaluator}
Existence studies employ both humans and LLMs as evaluators. Since the users of medical LLMs are humans, human evaluation can be effective and direct in assessing medical LLMs. Given the specialized nature of the medical field, existing work categorizes human evaluation into expert evaluation and lay user evaluation, reflecting the perspectives of physicians and general users, respectively. While human evaluation offers valuable insights into LLM performance, it is not without its limitations. Human evaluation can be time-consuming and resource-intensive, and the results can be susceptible to individual biases and expertise levels. To address these challenges, some researchers have explored employing advanced LLMs to replace human evaluation of medical LLMs. These advanced LLMs are trained on vast datasets and possess a knowledge base comparable to that of humans. Since the LLM evaluates the performance of medical LLMs from a human perspective, it is categorized under Human-Centric Evaluation.

\subsubsection{Evaluation Dimensions}
In Human-Centric Evaluation, the initial step involves identifying the evaluation dimensions for the model. Given the inconsistent terminology employed in previous studies of medical LLMs to describe these dimensions, we standardized the terminology and provided corresponding explanations. Tab. \ref{tab:eva} uses the normalized terms.
\par
\textbf{Doctor-like:} Exhibits a professional physician's communication style and tone, reflecting a comprehensive understanding of medical principles and clinical expertise. Capable of providing clear explanations of intricate medical concepts and offering well-founded diagnostic insights and recommendations.
\par
\textbf{Patient-friendly:} Interacts with users in a patient and empathetic manner, fostering a sense of understanding and support. Effectively simplifies complex medical concepts into easily comprehensible language, demonstrating compassion and care, alleviating user anxiety and enhancing trust and satisfaction.
\par
\textbf{Professional:} Exhibits a high degree of professionalism, characterized by comprehensive medical knowledge and sound clinical judgment. Provides treatment recommendations and prescriptions that are not only accurate and reliable but also conform to medical norms and ethics, ensuring the achievement of optimal patient outcomes.
\par
\textbf{Safe:} Responses are safe, harmless, and not geographically or racially discriminatory. Correctly address unhealthy and unsafe issues, provide accurate and reliable advice, and avoid misleading users or raising health risks. Always follow medical ethics, protect user privacy, and ensure the accuracy and safety of information delivery.
\par
\textbf{Fluent:} Responses should be fluent, coherent, and logical to ensure readability and comprehension. Accurate medical terminology should be employed without over-specialization, conveying the message professionally and understandably.
\par
\textbf{Proactive}: Capable of proactively asking users to add relevant information and ask targeted additional questions when faced with unclear descriptions of symptoms or lack of necessary information.
\par
\textbf{Helpful:} Capable of meeting the user's needs for medical consultation or the physician's clinical instructions to achieve users' expectations.
\par
Given the critical importance of safety in the medical field, the safety dimension must be assessed by medical experts. Other dimensions can be evaluated by LLMs, thereby increasing efficiency.
For non-professional users, the evaluation dimensions focus on patient-friendly and fluency. These two dimensions match the user's perspective and can measure the real feedback of the model from users in real scenarios.

\subsubsection{Protocols}
Human-centric evaluation comprises two primary protocols: individual evaluation and pairwise evaluation. 
\par
\textbf{Individual evaluation} involves a human evaluator or an LLM scoring the output based on predefined dimensions, with only a single question and its corresponding output visible. The scoring range is a predetermined interval, and the average score for each dimension is statistically calculated as the final result. This metric quantitatively reflects the model's overall performance in each dimension, identifying strengths and weaknesses and guiding future optimizations on weaker dimensions. 
\par
\textbf{Pairwise evaluation} involves a human evaluator or an LLM comparing the outputs of two models based on predefined dimensions and selecting the better-performing one. Evaluation metrics include win rate (the proportion of evaluated models outperforming the comparison model), tie rate (the proportion of the two that are equivalent), and loss rate (the proportion of evaluated models under-performing the comparison model). In addition, to minimize the positional bias existing in the LLM as a judge, the positions of the two model outputs are exchanged for another comparison.  Pairwise evaluation facilitates the rapid identification of the best-performing model from multiple candidates, serving as a foundation for model optimization and deployment.
\par
In addition to the above evaluation protocols, some open-source projects use \textbf{Case Study}, which relies entirely on subjective evaluation. It directly compares the responses of multiple medical LLMs to the same question, explicitly demonstrating the differences in medical knowledge and response style between them. It does not establish metrics for measuring model performance, relying entirely on the subjective perception of users.

\section{Challenges and Future Directions} \label{sec:challenges}
We analyze the current challenges of medical LLMs from the perspectives of data, methodology, and evaluation, and propose corresponding future research directions.
\subsection{Proactive Questioning Ability}
Compared to LLMs in other domains, medical LLMs must not only provide accurate responses to medical inquiries but also possess the capability to proactively guide users in describing their concerns clearly. Users often struggle to convey their conditions due to a lack of medical expertise. Thus, medical LLMs should ask users about unclear information and help them articulate the underlying causes of their illnesses. Additionally, users may provide incorrect descriptions of the symptoms of the disease, necessitating that medical LLMs correct these misunderstandings. Therefore, \textbf{the challenge is how to enhance the proactive questioning abilities of medical LLMs.} 
\par
To address this, Bianque \cite{Bianque} constructed a multi-round conversation corpus for model training, which balances the abilities to answer and ask questions, thereby improving the proactive questioning capability of medical LLMs at the corpus level. Currently, researchers may consider exploring the design of agent workflows. For instance, the medical counseling task can be decomposed into three components: (1) assessing the level of detail in the user's question, (2) identifying what information is missing, and (3) either answering the question or asking for more details from the user. Additionally, incorporating agent paradigms such as reflection and tool use into this workflow can enhance the credibility of responses generated by medical LLMs.

\subsection{Multi-organization joint training}
To enable medical LLMs to achieve capabilities comparable to those of medical professionals, it is essential to integrate more real medical data into these models to enhance their medical knowledge. Data from professional medical organizations tend to be more authentic and reliable than data sourced from the web. However, due to the sensitive nature of medical data and associated policies, such data is seldom publicly available \citep{gkoulalas2015medical}. Consequently, a significant \textbf{challenge is to ensure the secure utilization of medical data within organizations while safeguarding patient privacy.} One approach to addressing this issue is federated learning \citep{mcmahan2017communication}, a distributed training method that enables model parameters to be shared across multiple organizations without sharing the data itself. By storing data locally, federated learning preserves data privacy while allowing the model to learn from diverse data sources, thus enhancing its generalization and performance. Federated learning has been applied to medical images \citep{zhang2023fedbrain}, sensor data \citep{zhang2022federated}, and natural language text \citep{peng2024depth}, promoting the secure use of health data among medical organizations.
\par
Currently, federated learning has several shortcomings in the medical field that require further investigation. Differential privacy \citep{dwork2014algorithmic} can be integrated with federated learning to enhance data privacy protection, but this integration degrades model performance. How to balance privacy preservation and model performance in the medical LLMs training is a worthy research topic. Additionally, the quality and quantity of data varies among healthcare organizations, and how to evaluate the contribution of each organization is a crucial topic. The establishment of a standardized set of evaluation metrics to quantify the impact of different data sources could be considered. Finally, medical LLMs have a significantly higher number of parameters than pretrained language models. Researchers may consider optimizing communication and computational overhead during the transmission and aggregation of parameters in large-scale models. For example, explore how to combine parameter-efficient fine-tuning methods with federated learning.

\subsection{Personalized Diagnosis}
Before providing specific diagnoses and prescriptions, physicians typically inquire about patients' medical and allergy histories. However, the capabilities of medical LLMs rely mainly on training data that do not contain information about the currently attending patients. Consequently, proposed diagnoses and prescriptions may not be applicable to individual patients. Therefore, a significant \textbf{challenge is enabling medical LLMs to provide personalized services to patients or doctors using real-time information.} Retrieval-Augmented Generation (RAG) \citep{lewis2020retrieval} presents a feasible solution. This method first retrieves content relevant to the user query from external real-time sources (e.g., medical news, personal electronic medical records). It then integrates the user query with the retrieved content into a coherent prompt, and the model subsequently responds based on its own knowledge and the prompt \citep{gao2023retrieval}. 
Several medical LLM works have been developed to enhance responses using external knowledge, thereby improving performance and mitigating hallucination problems. For instance, ChatDoctor \citep{chatdoctor} employs online resources and custom offline medical databases as an external knowledge base to enhance accuracy in medical tasks through RAG. Moreover, JMLR \citep{JMLR} trains medical LLMs together with RAG to improve the model's ability to process medical knowledge while reducing computational resources.
\par
However, existing medical LLMs do not fully leverage the capabilities of RAG. The description or medical terminology provided by the user's limited medical knowledge may not accurately recall helpful content. Therefore, researchers may consider how to enrich or rewrite patient questions based on individual cases to improve information retrieval accuracy. Additionally, healthcare organizations typically maintain electronic medical records for their patients. Researchers can consider optimize the information retrieval embedding model for electronic health records to improve the accuracy and reduce the noise of retrieved information, thereby empowering medical large language models to enhance personalized diagnosis.

\subsection{Unified Evaluation Platform}
Currently, several benchmarks or platforms exist for evaluating general LLMs, such as GLUE \citep{wang2018glue}, SuperGLUE \citep{wang2019superglue}, CLUE \citep{xu2020clue}, and PromptCBLUE \citep{zhu2023promptcblue}. These platforms provide consistent evaluation criteria and standardized training and testing sets, facilitating fair comparisons and advancements among LLMs. However, existing medical LLMs are evaluated using inconsistent datasets and ambiguous evaluation tasks, resulting in comparisons and improvements between models becoming difficult.
Therefore, \textbf{it is particularly necessary to establish a unified evaluation platform focused on medical LLMs to foster the advancement of this field}. 
\par
Establishing a unified evaluation platform for medical LLMs necessitates attention to several critical considerations.
Firstly, the platform should incorporate benchmarks across various medical subdomains to ensure comprehensive evaluations of model performance across diverse areas. Secondly, the evaluation tasks should be clearly defined, encompassing medical QA, case diagnosis, medical reasoning, treatment recommendation, and other pertinent aspects to accurately reflect the model's practical applicability. Lastly, the evaluation platform should encompass various metrics and dimensions, including traditional measures such as accuracy and BLEU, alongside considerations of dimensions such as proactive, helpful, and safe.

\section{Conclusion}
This survey provides a guide on training customized medical LLMs based on general LLMs. We detail the training process at three levels: data, methodology, and evaluation. Specifically, we analyze the data level regarding  corpus sources and processing methods, the methodology level in terms of training paradigms, and the evaluation level from both machine and human perspectives. For each level, we provide corresponding recommendations for organizations that desire to develop a customized medical LLM. Finally, we present the challenges and future research directions for these three levels.

\printcredits

\bibliographystyle{model5-names}

\bibliography{cas-refs}

\end{document}